\documentclass[10pt,journal,compsoc]{IEEEtran}
\usepackage[nocompress]{cite}
\usepackage{array}
\usepackage{mdwmath}
\usepackage{mdwtab}
\usepackage{eqparbox}
\usepackage[hyphens,spaces,obeyspaces]{url}
\usepackage{amsmath}
\usepackage{float, xcolor}
\usepackage{graphicx}
\usepackage{multirow}
\usepackage{subfig}
\usepackage{booktabs}
\usepackage{amsmath,graphicx}
\usepackage{amsxtra}
\usepackage{threeparttable}
\usepackage[breaklinks=true]{hyperref}
\usepackage{bm}
\usepackage{enumitem}
\usepackage{ntheorem}
\usepackage{booktabs}
\usepackage{CJKutf8}
\usepackage{booktabs}
\usepackage{multirow}
\usepackage{subfig}

\ifCLASSOPTIONcompsoc
  \usepackage[nocompress]{cite}
\else
  \usepackage{cite}
\fi
\ifCLASSINFOpdf
\else
\fi
\begin{document}
%
\title{Speech Synthesis with Mixed Emotions}
%
%
%
%

\author{Kun Zhou,~\IEEEmembership{Student Member,~IEEE,}
        Berrak Sisman,~\IEEEmembership{Member,~IEEE,}
        Rajib Rana,~\IEEEmembership{Member,~IEEE,}\\
        Bj{\"o}rn W.\ Schuller,~\IEEEmembership{Fellow,~IEEE,}
        and Haizhou Li,~\IEEEmembership{Fellow,~IEEE}
\IEEEcompsocitemizethanks{

\IEEEcompsocthanksitem Kun Zhou is with the Department of Electrical and Computer Engineering, National University of Singapore, Singapore.
E-mail: zhoukun@u.nus.edu \protect
\IEEEcompsocthanksitem Berrak Sisman is with the Department of Electrical and Computer Engineering at the University of Texas at Dallas, United States. 
E-mail: berraksisman@u.nus.edu \protect
\IEEEcompsocthanksitem Rajib Rana is with the University of Southern Queensland, Australia. E-mail: Rajib.Rana@usq.edu.au\protect

\IEEEcompsocthanksitem Bj{\"o}rn W. Schuller is with GLAM -- the Group on Language, Audio, \& Music, Imperial College London, U.\,K., and the Chair of Embedded Intelligence for Health Care and Wellbeing, University of Augsburg, Germany. Email: bjoern.schuller@imperial.ac.uk\protect

\IEEEcompsocthanksitem Haizhou Li is with the Shenzhen Research Institute of Big Data, and the School of Data Science, The Chinese University of Hong Kong, Shenzhen, China; the Department of Electrical and Computer Engineering, National University of Singapore, Singapore; and the University of Bremen, Bremen, Germany. Email: haizhouli@cuhk.edu.cn\protect}}

\IEEEtitleabstractindextext{%
\begin{abstract}
Emotional speech synthesis aims to synthesize human voices with various emotional effects.
The current studies are mostly focused on imitating an averaged style belonging to a specific emotion type. In this paper, we seek to generate speech with a mixture of emotions at run-time.
We propose a novel formulation that measures the relative difference between the speech samples of different emotions. We then incorporate our formulation into a sequence-to-sequence emotional text-to-speech framework. 
During the training, the framework does not only explicitly characterize emotion styles but also explores the ordinal nature of emotions by quantifying the differences with other emotions. 
At run-time, we control the model to produce the desired emotion mixture by manually defining an emotion attribute vector. The objective and subjective evaluations have validated the effectiveness of the proposed framework. To our best knowledge, this research is the first study on modelling, synthesizing, and evaluating mixed emotions in speech. 
\end{abstract}

}

\maketitle

\IEEEdisplaynontitleabstractindextext

%
\IEEEpeerreviewmaketitle

\IEEEraisesectionheading{\section{Introduction}\label{sec:introduction}}
\IEEEPARstart{H}{umans} can feel multiple emotional states at the same time \cite{braniecka2014mixed}. Consider some bittersweet moments such as remembering a lost love with warmth or the first time leaving home for college, it is possible to experience the co-occurrence of different types of emotions - even two oppositely valenced emotions (e.\,g., happy and sad) \cite{larsen2014case,larsen2011further}. Emotional speech synthesis aims to add emotional effects to a synthesized voice \cite{schroder2001emotional}. Synthesizing mixed emotions will mark a milestone for achieving human-like emotions in speech synthesis, thus enabling a higher level of emotional intelligence in human-computer interaction \cite{pittermann2010handling,crumpton2016survey,rosenberg2021prosodic}.

Speech synthesis aims to generate human-like voices from input text \cite{tan2021survey,van2013progress,sisman2020overview}. With the advent of deep learning, the state-of-the-art speech synthesis systems \cite{sotelo2017char2wav,wang2017tacotron,ren2019fastspeech} are able to produce speech of high naturalness and intelligibility. However, most of them do not convey the omnipresent emotional contexts in human-human interaction \cite{ze2013statistical,yamagishi2009analysis,watts2016hmms}. The lack of expressiveness limits the emotional intelligence of current speech synthesis systems \cite{schuller2018age}. Emotional speech synthesis aims to fill this gap \cite{tokuda2002hmm,ohtani2015emotional,inoue2021model}.

Synthesizing a mixed emotional effect is a challenging task. 
One of the reasons is the subtle nature of human emotions~\cite{plutchik1991emotions}. Therefore, it is not straightforward to precisely characterize speech emotion.
Besides, speech emotion is inherently supra-segmental, complex with multiple acoustic cues such as timbre, pitch and rhythm \cite{xu2011speech,latorre2008multilevel}. 
Both spectral and prosodic variants need to be studied when modelling speech emotion. The early studies on emotional speech synthesis rely on statistical modelling of different speech parameters with hidden Markov models (HMM)~\cite{yamagishi2003modeling,eyben2012unsupervised} and Gaussian mixture model (GMM)~\cite{aihara2012gmm,kawanami2003gmm}. Deep neural networks (DNN)~\cite{lorenzo2018investigating, luo2017emotional} and deep bi-directional long-short-term memory network (DBLSTM)~\cite{ming2016deep,an2017emotional} represent the recent advances. The
end-to-end neural architecture~\cite{skerry2018towards,wu2019end} becomes popular because of its superior performance. We note that there are generally two types of methods in the literature to learn emotion information: one uses auxiliary emotion labels as the condition of the framework \cite{lee2017emotional,tits2019exploring}, and the other imitates the emotion style of the reference speech \cite{wang2018style,stanton2018predicting}. However, these methods learn the global temporal structure of speech emotion, resulting in a monotonous expressiveness in synthesized speech. In this way, these frameworks can only synthesize several emotion types  
exhibited in the database.
These disadvantages limit the flexibility and controllability of the above frameworks. For example, it is hard to synthesize mixed emotional effects with existing emotional speech synthesis frameworks.


For the first time, we study the modelling of mixed emotions in speech synthesis. 
In psychology, there have been studies~\cite{kreibig2017understanding,berrios2015eliciting}  to understand the paradigms and measures of mixed emotions. However, the study of mixed emotions in speech synthesis is not given attention yet, where 
there exist two main research problems: (1) how to characterize and quantify the mixture of speech emotions, and (2) how to evaluate the synthesized speech. In this article, we will address these two challenges. 


The main contributions of this article are listed as follows:
\begin{itemize}

    \item For the first time, we study the modelling of mixed emotions for speech synthesis, which brings us a step closer to achieving emotional intelligence; 
    
    
    \item
   {We introduce a novel scheme to measure the relative difference between emotion categories, with which the emotional text-to-speech framework learns to quantify the differences between the emotion styles of speech samples during the training.
    At run-time, we control the model to produce the desired emotion mixture by manually defining an emotion attribute vector;}
    
    \item We carefully devise objective and subjective evaluations to confirm the effectiveness of the proposed framework and the emotional expressiveness of the speech.

\end{itemize}


This paper is organized as follows: In Section \ref{sec:related work}, we motivate our study by introducing the background and related work. In Section \ref{sec: main}, we present the details of our proposed framework, and we introduce our experiments in Section~\ref{sec: exp}. We provide further investigations in Section \ref{sec: further}. The study is concluded in Section \ref{sec: concludes}.

\section{Background and Related Work}
This work is built on several previous studies on the characterization of emotions, sequence-to-sequence emotion modelling for speech synthesis and controllable emotional speech synthesis. We briefly introduce the related studies to set the stage for our research and rationalize the novelty of our contributions.

\label{sec:related work}

\subsection{Characterization of Emotions}
Understanding human emotions (e.\,g., their nature and functions) has been gaining lots of attention in psychology \cite{niedenthal2012social,hogan2003mind,ekman1994nature}. This study is inspired by several previous research, including the theory of the emotion wheel and the ordinal nature of emotions. 

   \subsubsection{Theory of the Emotion Wheel}

\begin{figure}[t]
    \centering
    \includegraphics[width=0.45\textwidth]{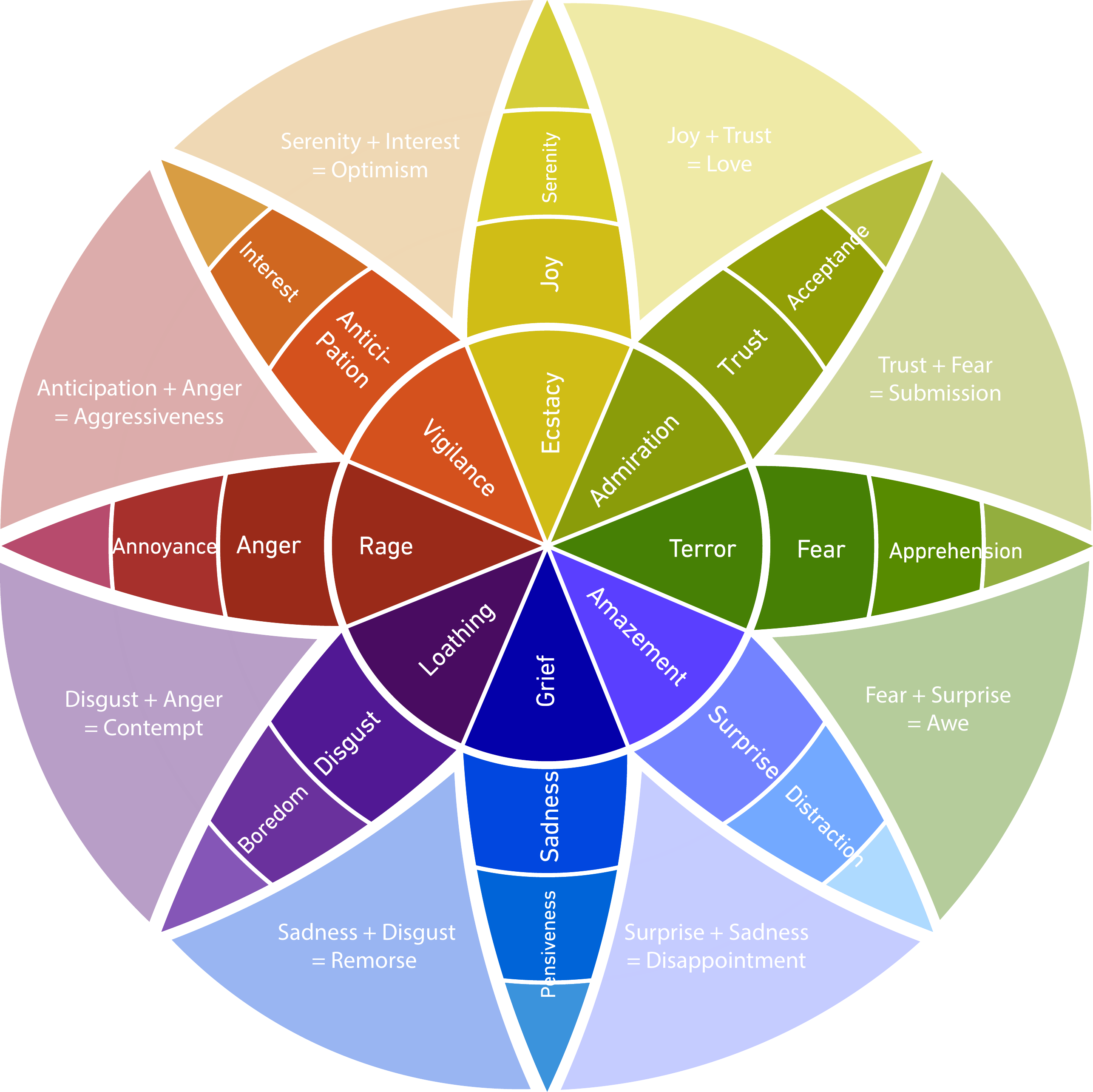}
    \caption{An illustration of the theory of the emotion wheel \cite{plutchik2013theories}, where all emotions occur as the mixed or derivative states of eight primary emotions.    }
    \label{fig:emotion_wheel}
\end{figure}

Humans can experience around $34,000$ different emotions \cite{plutchik2001nature}. While it is hard to understand all these distinct emotions, Plutchik proposed $8$ primary emotions: anger, fear, sadness, disgust, surprise, anticipation, trust and joy, and arranged them in an emotion wheel \cite{plutchik2013theories} as shown in Figure \ref{fig:emotion_wheel}. All other emotions can be regarded as mixed or derivative states of these primary emotions \cite{plutchik2013theories}. According to the theory of the emotion wheel, the changes in intensity could produce the diverse amount of emotions we can feel. Besides, the adding up of primary emotions could produce new emotion types.
For example, delight can be produced by combining joy and surprise \cite{cross2016changing}.

Despite these efforts in psychology, there is almost no attempt to model the mixed emotions in the literature of speech synthesis. Inspired by the theory of the emotion wheel,  we believe it is possible to combine different primary emotions and synthesize mixed emotions in speech. This technique will also allow us to create new emotion types that are hard to collect in real life, which could help us better mimic human emotions and further enhance the engagement in human-robot interaction.

\subsubsection{The Ordinal Nature of Emotions}

Emotions are intrinsically relative, and their annotations and analysis should follow the ordinal path \cite{yannakakis2017ordinal,yannakakis2018ordinal}. Instead of assigning an absolute score or an emotion category, ordinal methods characterize emotions through comparative assessments (e.\,g., is sentence one \textit{happier} than sentence two?). Ordinal methods have shown remarkable performance, especially in speech emotion recognition \cite{harvill2021quantifying,cao2015speaker,harvill2019retrieving}. 

The key idea of ordinal methods is to learn a ranking according to the given criterion. An example is preference learning \cite{furnkranz2003pairwise}, where the task is to establish preferences between samples. Once the preferences are established, ranking samples \cite{lotfian2016retrieving,parthasarathy2016using,martinez2014don} is straightforward. Other rank-based methods \cite{yannakakis2015grounding,huang2018speech,feng2020siamese} also show the effectiveness of modelling the affect for speech emotion recognition. 
As for emotional speech synthesis, researchers also explore the ordinal nature of emotions to model the emotion intensity \cite{zhu2019controlling,lei2021fine,zhou2022emotion,lei2022msemotts}, where the intensity of an emotion is treated as the relative difference between neutral and emotional samples. Inspired by the previous studies, we aim to study rank-based methods to quantify the relative differences between the speech samples from different emotion categories, which we discuss later. 

\subsection{Sequence-to-Sequence Emotion Modelling for Speech Synthesis}

The sequence-to-sequence model with attention mechanism was first studied in machine translation \cite{bahdanau2015neural} and later on found effective in speech synthesis \cite{kyle2017char2wav,wang2017tacotron}. We consider that sequence-to-sequence models are suitable for modelling speech emotion. Sequence-to-sequence models are more effective in modelling the long-term dependencies at different temporal levels such as word, phrase and utterance \cite{schuller2020review}.
By learning attention alignment, 
sequence-to-sequence models can capture the dynamic prosodic variants within an utterance \cite{zhou2021limited}. They also allow for the prediction of the speech duration at run-time, which is a critical prosodic factor of the speech emotion \cite{wu2010acoustic}.

There are generally two types of methods in the literature to model speech emotions: 1) explicit label-based and 2) reference-based approaches. Next, we will briefly introduce these two approaches in sequence-to-sequence modelling.  

\subsubsection{Learn to Associate with Explicit Labels}
It is the most straightforward to characterize emotion by using explicit emotion labels~\cite{lee2017emotional,tits2019exploring}, where the model learns to associate labels with emotion styles. In \cite{lee2017emotional}, an emotion label vector is taken by the attention-based decoder to produce the desired emotion. In \cite{tits2019exploring}, a low-resourced emotional text-to-speech is built using model adaptation with a few emotion labels. In addition to the explicit labels of discrete emotion categories, there are attempts to condition the decoder with continuous variables \cite{rabiee2019adjusting}.

\subsubsection{Learn to Imitate a Reference}
Another approach is to use a style encoder to imitate and transplant the reference style \cite{skerry2018towards}. Global style token (GST) \cite{wang2018style} is an example to learn style embeddings from the reference audio in an unsupervised manner. Some studies incorporate additional emotion recognition loss \cite{wu2019end,cai2021emotion}, perceptual loss \cite{li2021controllable,zhou2022emotion} or adversarial training \cite{ma2018neural} to help with the emotion rendering. Other studies \cite{cornille2022interactive, klimkov19_interspeech,li21r_interspeech,zhang2020learning} replace the global style embedding with phoneme or segmental level prosody embedding to capture multi-scale emotion variants. 
Similar approaches have also been applied to emotional voice conversion research. In \cite{zhou21b_interspeech}, the style encoder further acts as the emotion encoder to learn actual emotion information through a two-stage training. In \cite{choi2021sequence}, a speaker encoder is further introduced to preserve the speaker information. 

These successful attempts motivate us to leverage the sequence-to-sequence mechanism to enable emotion modelling for speech synthesis.

\subsection{Controllable Emotional Speech Synthesis}

Speech emotion is often manifested in various prosody aspects \cite{mozziconacci2002prosody}. Emotion rendering can be controlled by modifying different prosodic cues. Current studies \cite{lee2019robust,tan21_interspeech} mainly focus on designing the prosody embedding as a control vector that is derived from a representation learning framework. For example, style tokens \cite{wang2018style} are designed to represent high-level styles such as speaker style, pitch range and speaking rate. Emotion rendering can be controlled by choosing specific tokens. Recent attempts \cite{sun2020fully,sun2020generating} study a way to include a hierarchical, fine-grained prosody representation into the style token-based diagram \cite{wang2018style}. 
Some other studies also use variational autoencoders (VAE) \cite{kingma2013auto} to control the speech style by learning, scaling or combining disentangled representations \cite{zhang2019learning,kenter2019chive}.

Recently, emotion intensity control has attracted much attention in emotional speech synthesis. 
Emotion intensity is considered to be correlated with all the acoustic cues that contribute to speech emotion \cite{frijda1992complexity}, which makes itself even more subjective and challenging to model. Some studies use auxiliary features such as a state of voiced, unvoiced and silence (VUS) \cite{matsumoto20_interspeech}, attention weights or a saliency map \cite{schnell11improving} to control the emotion intensity. Other studies manipulate the internal emotion representations through interpolation \cite{um2020emotional}, scaling \cite{choi2021sequence} or distance-based quantization \cite{im2022emoq}. In \cite{zhu2019controlling,lei2021fine,zhou2022emotion,lei2022msemotts}, relative attributes are introduced to learn a more interpretable representation of emotion intensity. However, none of these frameworks studied the correlation and interplay between different emotions. This contribution aims to fill this research gap. 

\begin{figure}[t]
    \centering
    \includegraphics[width=0.5\textwidth]{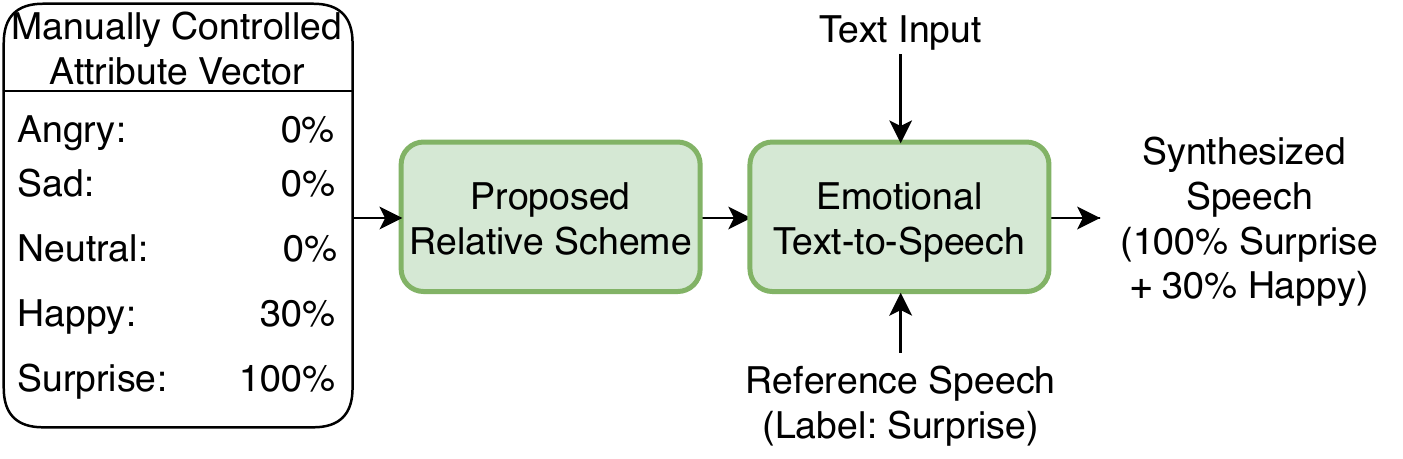}
  \caption{Block diagram of our proposed relative scheme applied to emotional text-to-speech at run-time.}
    \label{fig:diagram}
\end{figure}

\subsection{Summary of Research Gap}

We briefly summarize the gaps in the current literature on speech synthesis that we aim to address in this study:
\begin{itemize}
    \item The synthesis of mixed emotions has not been studied in speech synthesis, which limits the capability of current systems to imitate  human emotions;
    
    \item Despite much progress in psychology, it is still challenging to characterize and quantify the mixture of emotions in speech;
    
    \item Current evaluation methods are inadequate to assess mixed emotional effects. The rethinking of the current evaluation for mixed emotions is needed.
    
\end{itemize}
This study is a departure from the current studies on emotional speech synthesis. We seek to display the possibilities to synthesize mixed emotions that are subtle but do exist in our real life.

\begin{figure*}[t]
    \centering
    \includegraphics[width=0.7\textwidth]{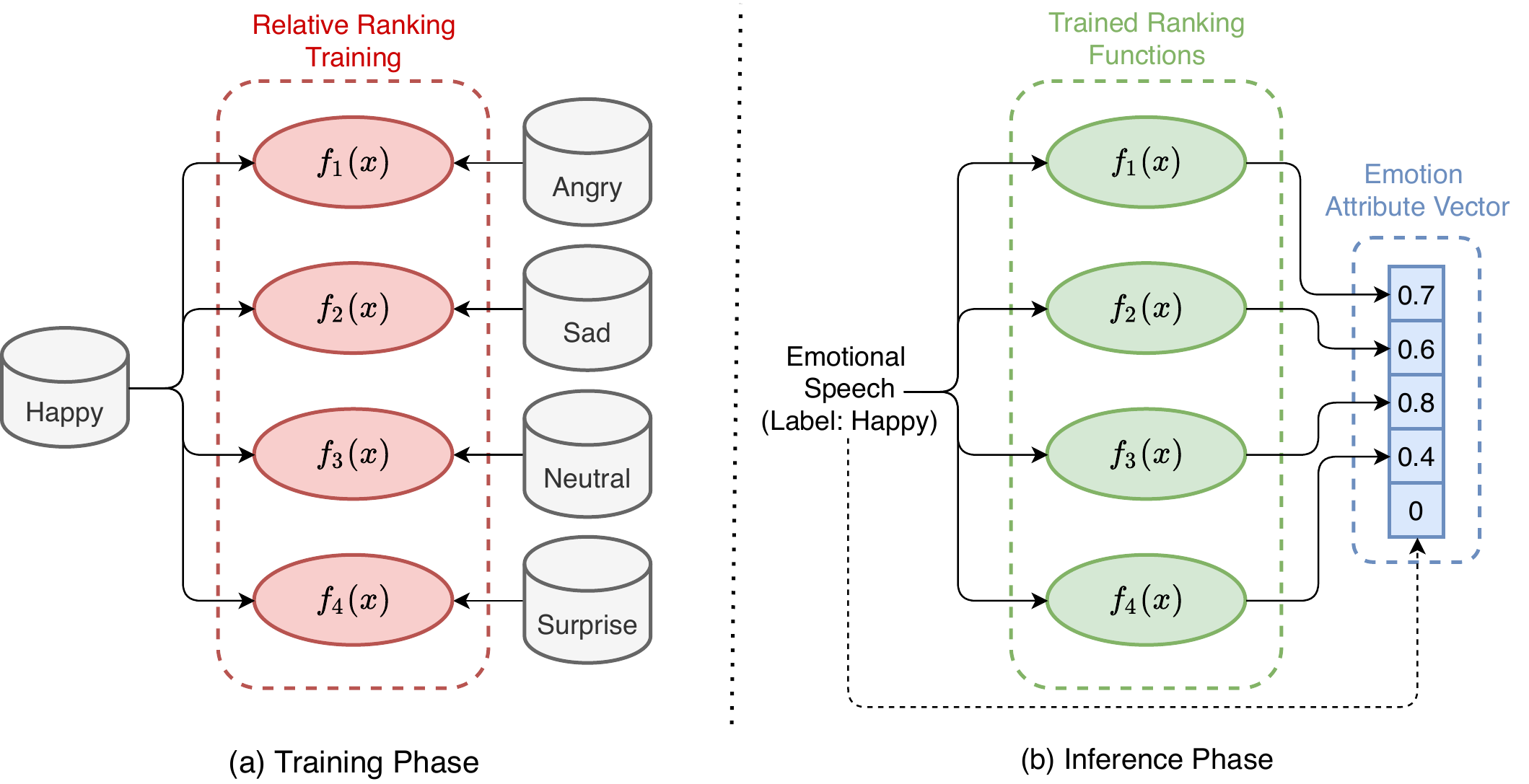}
    \caption{The illustration of the proposed relative scheme at (a) training and (b) run-time phase. A relative ranking function is trained between each emotion pair and automatically predicts an emotion attribute at run-time. A smaller emotion attribute value represents a similar emotional style between the pairs. All the emotion attributes form an emotion attribute vector.   }
    \label{fig:scheme}
\end{figure*}

\section{Mixed Emotion Modelling and Synthesis }
\label{sec: main}

We propose a novel relative scheme that allows for manually manipulating the synthesized emotion, i.e. mixing multiple different emotion styles. As shown in Figure \ref{fig:diagram}, the proposed scheme allows for flexible control of the extent of each contributing emotion in the speech. At run-time, the framework transfers the reference emotion into a new utterance with the text input, also known as emotional text-to-speech.

We first describe our method of characterizing mixed emotions in speech and highlight our contributions to designing a novel relative scheme. Then, we present the details of the sequence-to-sequence emotion training with the proposed relative scheme. Lastly, we show the flexible control of the proposed framework for synthesizing mixed emotions. 

 \subsection{Characterization of Mixed Emotions in Speech}
 \label{characterization}
Emotion can be characterized with either categorical \cite{whissell1989dictionary,ekman1992argument} or dimensional representations \cite{russell1980circumplex,schroder2006expressing}. 
With designated emotion labels, the emotion category approach is the most straightforward way to represent emotions. However, such representation ignores the subtle variations of emotions. 
Another approach seeks to model the physical properties of speech emotion with dimensional representations. An example is Russell's circumplex model \cite{russell1980circumplex}, where emotions are distributed in a two-dimensional circular space, containing arousal and valence dimensions.

One of the most straightforward ways to characterize mixed emotions is to inject different emotion styles into a continuous space. Mixed emotions could be synthesized by adjusting each dimension carefully. However, only a few emotional speech databases \cite{busso2008iemocap,busso2016msp} provide such annotations. These dimensional annotations are subjective and expensive to collect. Therefore, we only utilize discrete emotion labels available in most databases. We first make an assumption based on the theory of the emotion wheel \cite{plutchik2013theories}:
Mixed emotions are characterized by combinations, mixtures, or compounds of primary emotions. While it is not straightforward to add up emotions, we explore the ordinal nature of emotions instead.

We propose a rank-based relative scheme to quantify the relative difference between speech recordings with different emotion types. Mixed emotions can be characterized by adjusting the relative difference with other emotion types. The relative difference value can also quantify the level of engagement of each emotion. We introduce our design of a novel relative scheme next.

\begin{figure}
    \centering
    \includegraphics[width=0.5\textwidth]{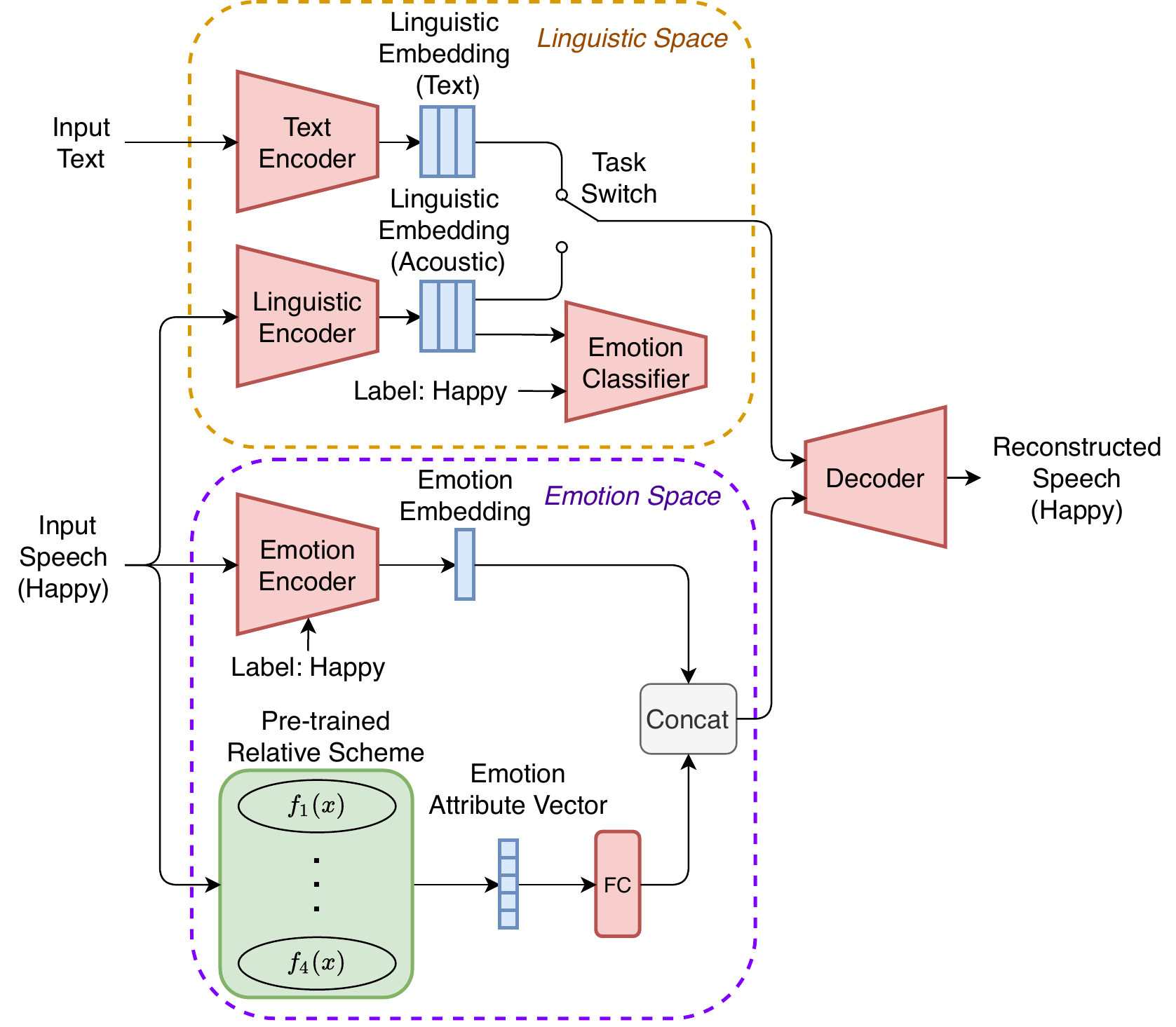}
    \caption{The training diagram of the proposed framework. The pre-trained relative scheme learns to generate an emotion attribute vector that measures the relative difference between the input emotion style ('Happy') and other primary emotion styles ('Angry', 'Sad', 'Surprise' and 'Neutral').}
    \label{fig:train}
\end{figure}
\subsection{Design of a Novel Relative Scheme}
\label{sec:design}

One of the challenges of synthesizing mixed emotions is quantifying the association or the interplay between different emotions. Inspired by the ordinal nature of emotions, we propose a novel relative scheme to address this challenge.
We first make two assumptions according to the theory of the emotion wheel: (1)
all emotions are related to some extent; (2) each emotion has stereotypical styles.
In our proposal, we not only characterize the identifiable styles of each emotion but also seek to quantify the similarity between different emotion styles.

We study a rank-based method to measure the relative difference between emotion categories, which can offer more informative descriptions and thus be closer to human supervision \cite{parikh2011relative}. In computer vision, the relative attribute \cite{parikh2011relative} represents an effective way to model the relative difference between two categories of data. Inspired by the success in various computer vision tasks \cite{kovashka2012whittlesearch,zhang2015robust,fan2013relative}, we believe relative attributes bridge between the low-level features and high-level semantic meanings, which allows us to model the relative difference between emotions only with discrete emotion labels. 
 In this way, we regard the identifiable emotion style as an attribute of speech data, which can be represented with a rich set of emotion-related acoustic features. The relative difference of the emotion styles can be modelled as a relative attribute, which is called "\textit{emotion attribute}" in this article. The emotion attribute can be learned through a max-margin optimization problem as explained below:

Given a training set $T = \{\mathbf{x}_n\}$, where ${\mathbf{x}_n}$ is the acoustic features of the $n^{th}$ training sample, and $T = A \cup B$, where $A$ and $B$ are two different emotion sets, we aim to learn a ranking function given as below:
\begin{equation}
    f(x_n) = \mathbf{W}\mathbf{x}_n, 
\end{equation}
where $\mathbf{W}$ is a weighting matrix indicating the difference in emotion styles. 
According to hypotheses (1) and (2), we propose the following constraints:
\begin{align}
    \forall   \mathbf{x}_i\in A, \forall   \mathbf{x}_j\in B: \mathbf{W}\mathbf{x}_i > \mathbf{W}\mathbf{x}_j\\
    \forall   (\mathbf{x}_i, \mathbf{x}_j)\in A, \forall  (\mathbf{x}_i, \mathbf{x}_j)\in B: \mathbf{W}\mathbf{x}_i = \mathbf{W}\mathbf{x}_j, 
\end{align}

The weighting matrix $\mathbf{W}$ is estimated by solving the following problem similar to that of a support vector machine \cite{chapelle2007training}:

\begin{align}
    \min_{\mathbf{W}} (\frac{1}{2} \parallel \mathbf{W} \parallel_2^2 + C(\sum \xi_{i,j}^2 + \sum \gamma_{i,j}^2))\\
    \text{ s.t. }           \mathbf{W}(\mathbf{x}_i - \mathbf{x}_j) \geq 1 - \xi_{i,j}; \forall  \mathbf{x}_i\in A, \forall   \mathbf{x}_j\in B\\
    |\mathbf{W}(\mathbf{x}_i-\mathbf{x}_j)|\leq \gamma_{i,j}; \forall   (\mathbf{x}_i, \mathbf{x}_j)\in A, \forall  (\mathbf{x}_i, \mathbf{x}_j)\in B\\
    \xi_{i,j} \geq 0; \gamma_{i,j}\geq 0, 
\end{align}
where $C$ is the trade-off between the margin and the size of slack variables $\xi_{i,j}$ and $\gamma_{i,j}$. 

Through Eq.\ (4) -- (7), we learn a wide-margin ranking function that enforces the ordering on each training point. As shown in Figure \ref{fig:scheme}(a), we train a relative ranking function $f(x)$ between each emotion pair. At the inference phase, the trained function can estimate an emotion attribute of unseen data, as shown in Figure \ref{fig:scheme}(b). In practice, each emotion attribute value is normalized to $[0,1]$, where a smaller value indicates a similar emotional style. All the normalized emotion attributes form an emotion attribute vector. The emotion attribute vector bridges the discrete primary emotion labels and is further incorporated in sequence-to-sequence emotion training.

\begin{figure}[t]
    \centering
    \includegraphics[width=0.5\textwidth]{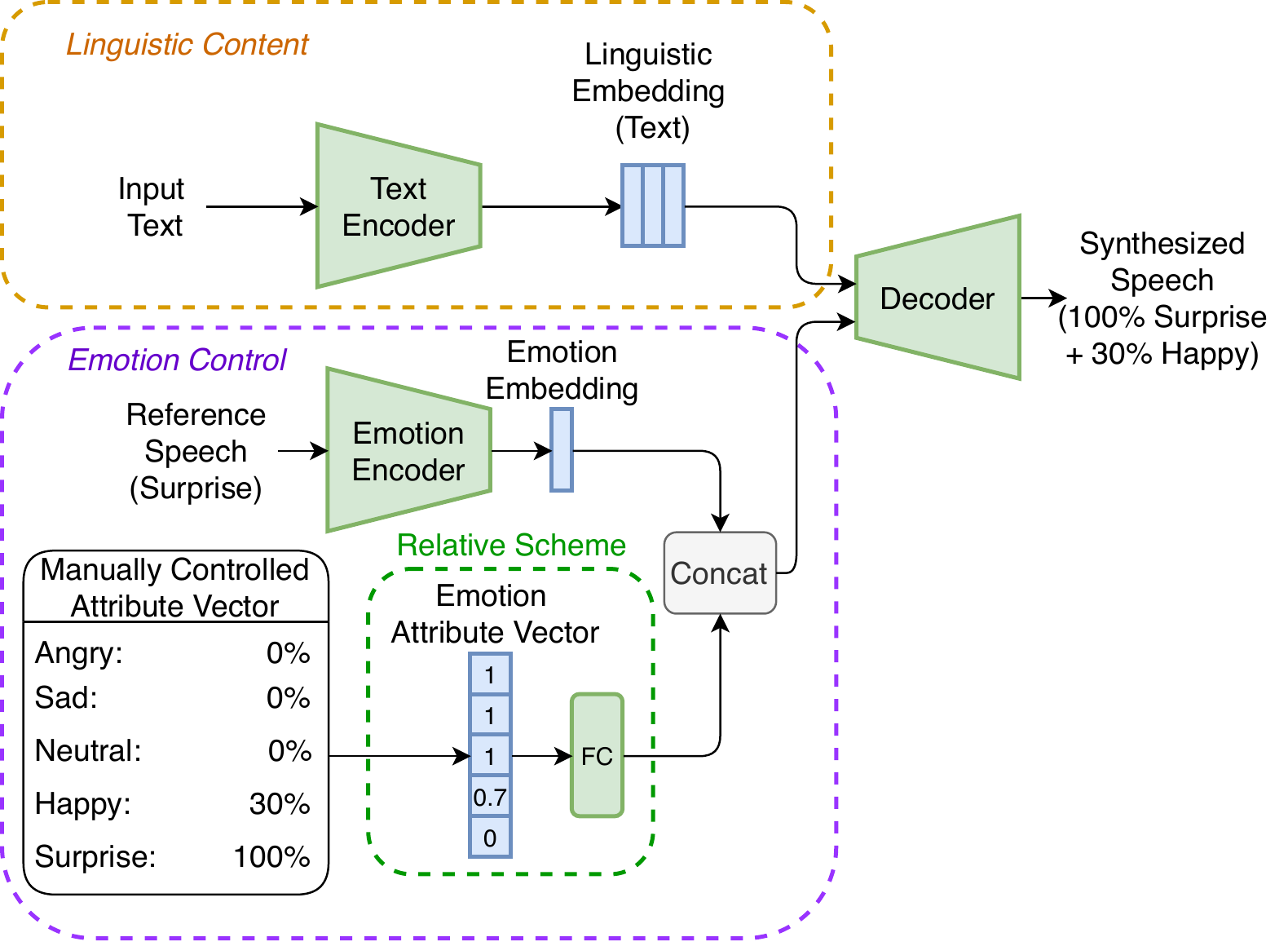}
    \caption{The run-time diagram of the proposed emotional text-to-speech framework. The emotion rendering can be manually controlled via the relative scheme. By assigning the appropriate percentage to the attribute vector, we produce a target emotion mixture.} 
    \label{fig:run-time}
\end{figure}

\subsection{Training Strategy}

{We adopt an emotional text-to-speech framework with the joint training of voice conversion as in \cite{zhou21b_interspeech}. As both text-to-speech and voice conversion share a common goal of generating realistic speech from the internal representations, the joint training was shown effective~\cite{zhang2019joint, zhang2019non,zhang2021transfer,polyak2019attention}. The text-to-speech task could benefit from the phone-embedding vectors \cite{ping2017deep,li2018close}, or the prosody style introduced by a reference encoder \cite{skerry2018towards}. A shared decoder between text-to-speech and voice conversion contributes to a robust decoding process \cite{luong2019bootstrapping,huang2020voice,luong2020nautilus}.  }

The overall emotional text-to-speech framework is an encoder-decoder model that is trained as a sequence-to-sequence system, as shown in Figure \ref{fig:train}, where the text encoder and linguistic encoder generate an embedding sequence for the input, while the emotion encoder generates one embedding that encapsulates the whole reference speech sample. 

Given the text or speech as input, the text and the linguistic encoder learn to predict the linguistic embedding from the text or speech, respectively. The decoder takes the linguistic embedding from the text or speech in an alternative manner, depending on whether the epoch number is odd or even. Similar to \cite{zhang2019non}, a contrastive loss is used to ensure the similarity between these two types of linguistic embeddings. The adversarial training strategy with an emotion classifier is employed on the acoustic linguistic embedding to eliminate the residual emotion information. 

An emotion encoder is used to extract an emotion embedding vector from the input speech under the supervision of an emotion label. 
Meanwhile, an emotion attribute vector is generated by the pre-trained relative scheme described in Section \ref{sec:design}, and then produced by a fully connected (FC) layer, resulting in a relative embedding. The emotion embedding describes the emotion styles of the input speech, while the emotion attribute vector indicates the difference between the input emotion style and other emotion styles. Finally, the decoder learns to reconstruct the input emotion style from a combination of emotion and relative embeddings. 

The whole training procedure can be viewed as a recognition-synthesis process at the sequence level. Our proposed framework does not only learn the abundant emotion variance that is exhibited in a database but also the correlation or association across different emotion categories. It allows us to explicitly adjust the difference level at run-time and further enables 
mixed emotion synthesis and the flexible control of emotion rendering at the same time, which will be discussed next.

\subsection{Control of Emotion Rendering}

We illustrate our proposed emotional text-to-speech framework in Figure \ref{fig:run-time}, which renders controllable emotional speech at run-time. The framework consists of three main modules, the content encoder, the emotion controller, and the decoder. 

The text encoder projects  the linguistic information from the input text into an internal representation. 
The emotion encoder captures the emotion style in an embedding from the reference speech, while the relative scheme further introduces the characteristics of other emotion types with a manually assigned attribute vector. By varying the percentage for each primary emotion in the attribute vector, we can easily synthesize the desired emotional effects and control the emotion rendering in synthesized speech.

\section{Experiments and Evaluations}
\label{sec: exp}

In this section, we report our experimental settings and results. As shown in Table \ref{tab: mix}, for all the experiments, we synthesize mixed emotional effects by mixing a primary emotion (\textit{Surprise}) with three reference emotions (\textit{Happy}, \textit{Angry} and \textit{Sad}) respectively. We expect to synthesize mixed emotional effects similar with the secondary emotions such as \textit{Delight}, \textit{Outrage} and \textit{Disappointment}, respectively. We choose these three combinations because they are thought to be easier to perceive for the listeners and have been studied in psychology \cite{braniecka2014mixed,plutchik2013theories}. 

Since this contribution serves as a pioneer in related fields, there is no literature or reference method before this study, to our best knowledge. Therefore, we could not include any baselines in our experiments. Instead, we adopt objective and subjective metrics widely used in previous literature and carefully design evaluation methods to show the effectiveness of our proposal. We have made the source codes and speech demos available to the public\footnote{\textbf{Codes \& Speech Demos}: \href{https://kunzhou9646.github.io/Mixed_Emotions_Demo/}{\nolinkurl{https://kunzhou9646.github.io/Mixed_Emotions_Demo/}}}. We encourage readers to listen to the speech samples on our demo website to 
best 
understand this work.

\subsection{Experimental Setup}

We use acoustic features and phoneme sequences as inputs to the proposed framework during the training. The acoustic features are 80-dimensional  logarithm Mel-spectrograms extracted every $12.5$\,ms with a frame size of $50$\,ms for short-time Fourier transform (STFT). We convert text to phoneme with the Festival \cite{black2001festival} G2P tool to serve as the input to the text encoder. At run-time, we synthesize emotional speech from the text input.

\subsubsection{Network Configuration}

Our proposed framework can be regarded as a sequence-level recognition-synthesis structure similar to that of \cite{zhang2019non,zhang20d_interspeech}. Both the linguistic encoder and the decoder have a sequence-to-sequence encoder-decoder structure.
The linguistic encoder consists of an encoder, a 2-layer 256-cell BLSTM and a decoder, a 1-layer 512-cell BLSTM with an attention layer followed by a full-connected (FC) layer with an output channel of $512$. 
The decoder has the same model architecture as that of Tacotron \cite{wang2017tacotron}.

The text encoder is a 3-layer 1D CNN with a kernel size of $5$ and a channel number of $512$. The text encoder is followed by a 1-layer of 256-cell BLSTM and an FC layer with an output channel number of $512$. 
The style encoder is a 2-layer 128-cell BLSTM followed by an FC layer with an output channel number of $64$. The classifier is a 4-layer FC with channel numbers of $\{$512, 512, 512, 5$\}$.
\begin{table}[t]
\centering
\caption{Our experimental settings of one primary emotion (A), three reference emotions (B) and the expected mixed emotional effects (A+B).}
\scalebox{0.9}{
\begin{tabular}{c|c|c}
\toprule
\textbf{Primary Emotion (A)} & \textbf{Reference Emotion (B)} & \textbf{Mixed Effects (A+B)} \\ \midrule
Surprise             & Happy          & Delight            \\ 
Surprise             & Angry          & Outrage             \\ 
Surprise             & Sad          & Disappointment      \\ \bottomrule
\end{tabular}}
\label{tab: mix}
\end{table}

 \begin{figure*}[t]
\centering
\subfloat[Mixing Surprise (100\%) with different percentages of Angry (0\%, 30\%, 60\%, 90\%)]
{
\begin{minipage}[c]{0.33\linewidth}
\centering
\includegraphics[width=1\textwidth]{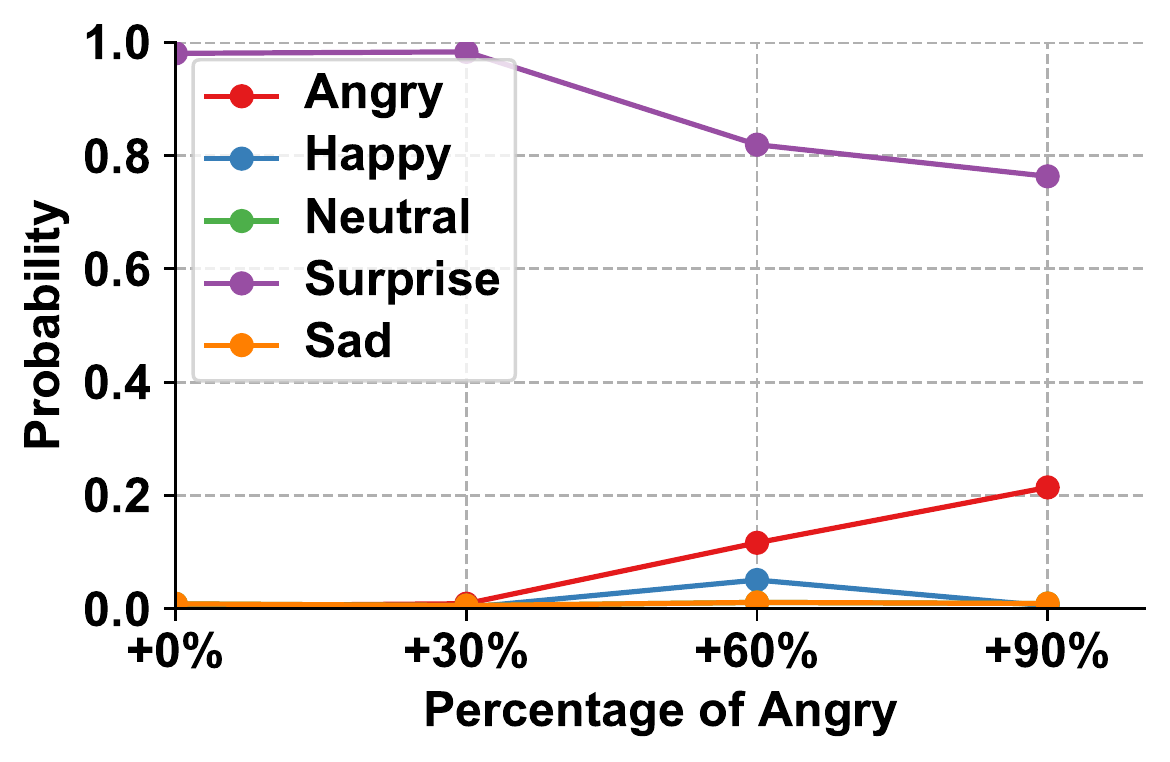}
\end{minipage}%
}
\subfloat[Mixing Surprise (100\%) with different percentages of Happy (0\%, 30\%, 60\%, 90\%)]
{
\begin{minipage}[c]{0.33\linewidth}
\centering
\includegraphics[width=1\textwidth]{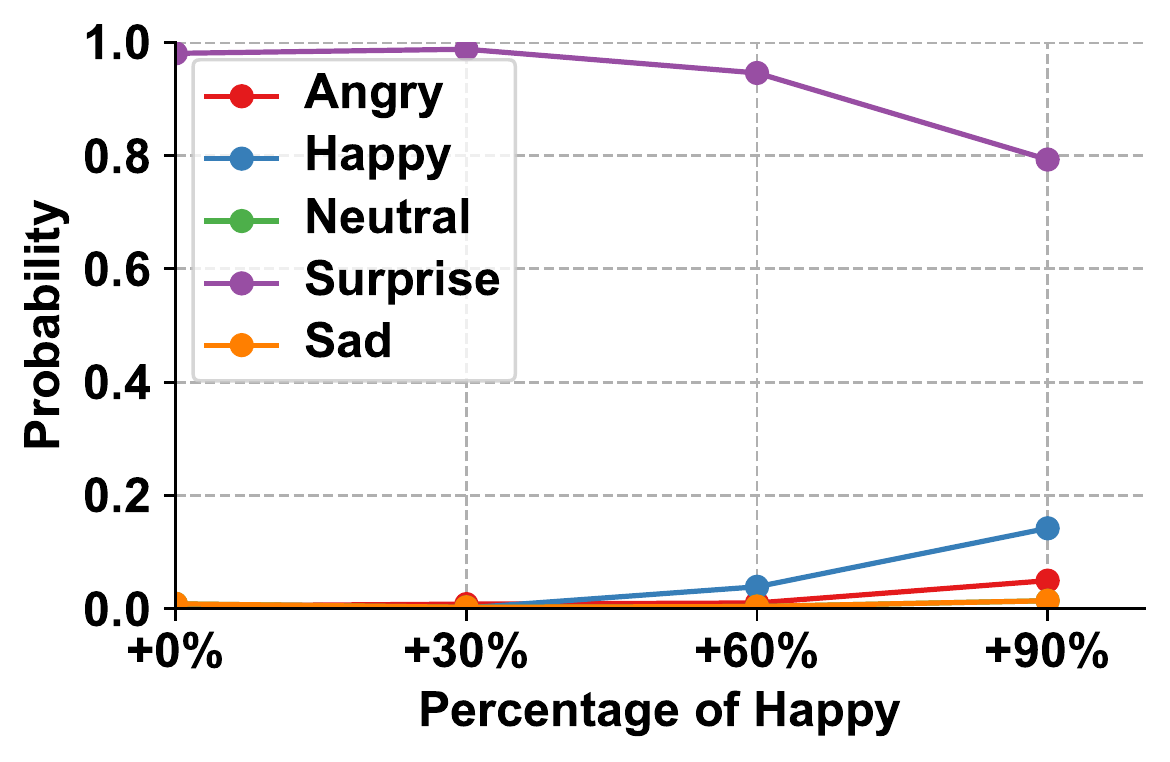}
\end{minipage}
}
\subfloat[Mixing Surprise  (100\%) with different percentages of Sad (0\%, 30\%, 60\%, 90\%)]
{
\begin{minipage}[c]{0.33\linewidth}
\centering
\includegraphics[width=1\textwidth]{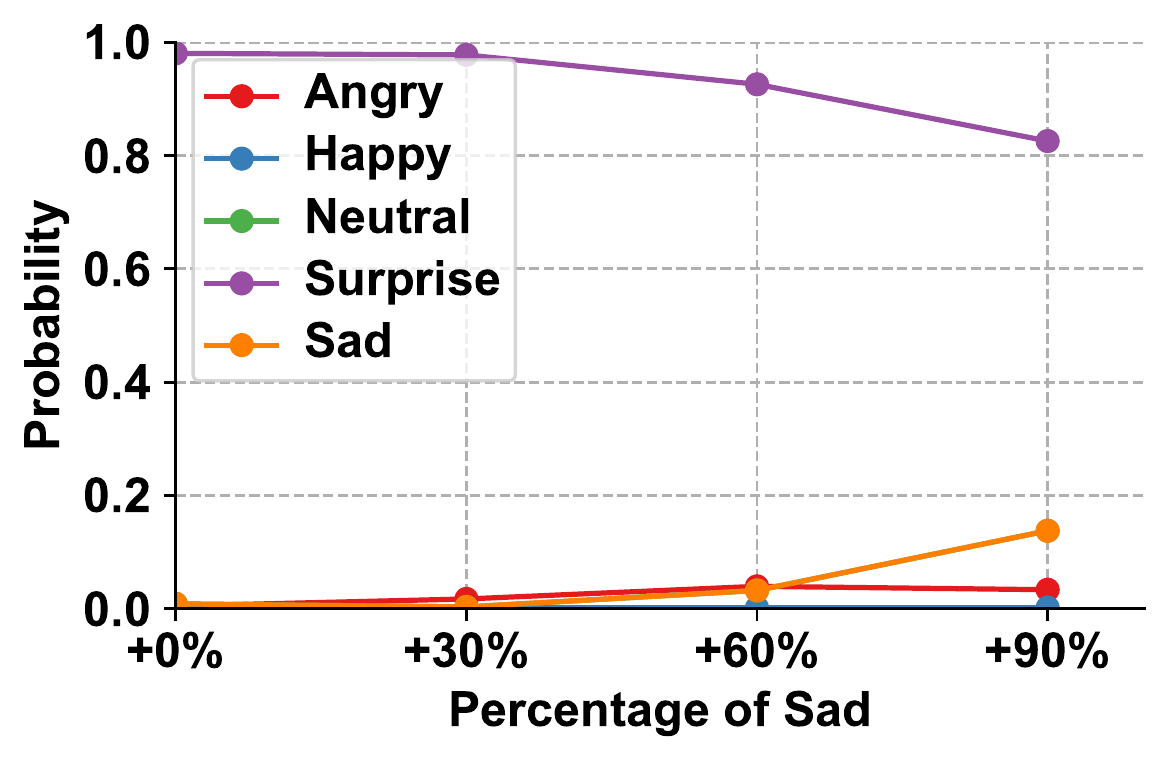}
\end{minipage}
}
\centering
\caption{Classification probabilities derived from the pre-trained SER model for a male speaker ('0013') from the ESD dataset.  Each point represents an averaged probability value of 20 utterances with mixed emotions.}
\label{fig:ser_male}
\end{figure*}
\begin{figure*}[t]
\centering
\subfloat[Mixing Surprise (100\%) with different percentages of Angry (0\%, 30\%, 60\%, 90\%)]
{
\begin{minipage}[c]{0.33\linewidth}
\centering
\includegraphics[width=1\textwidth]{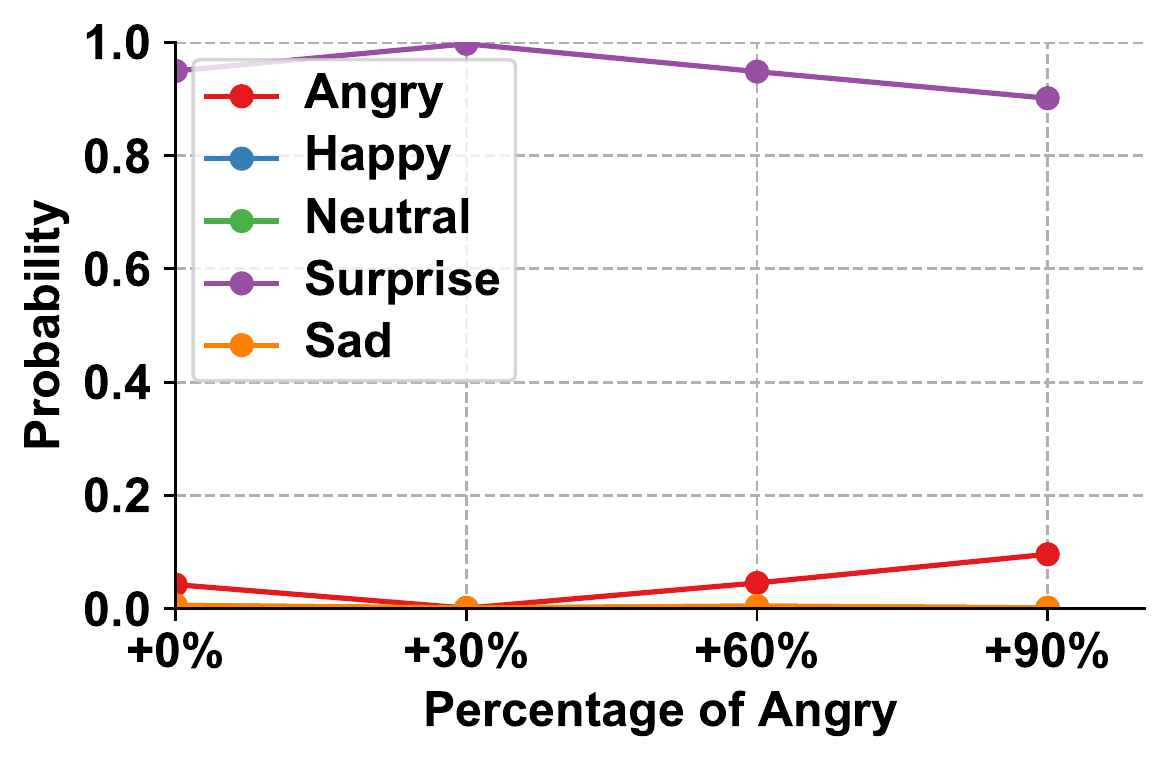}
\end{minipage}%
}
\subfloat[Mixing Surprise (100\%) with different percentages of Happy (0\%, 30\%, 60\%, 90\%)]
{
\begin{minipage}[c]{0.33\linewidth}
\centering
\includegraphics[width=1\textwidth]{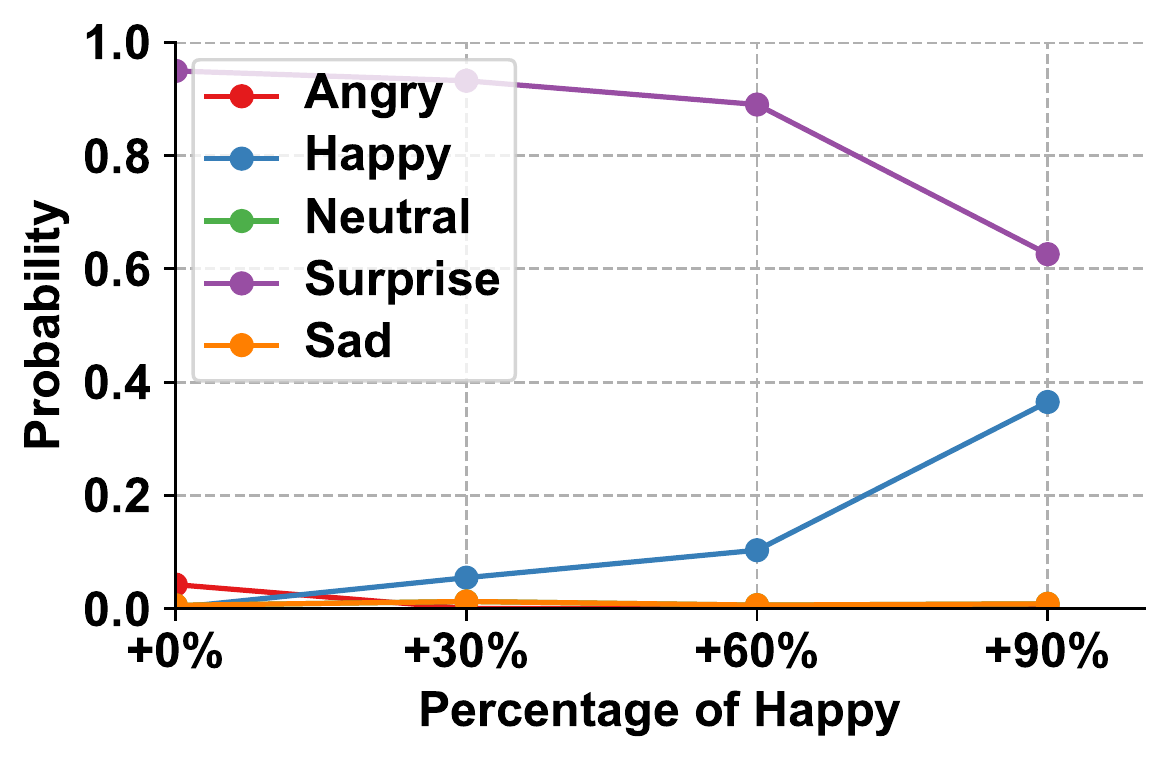}
\end{minipage}
}
\subfloat[Mixing Surprise (100\%) with different percentages of Sad (0\%, 30\%, 60\%, 90\%)]
{
\begin{minipage}[c]{0.33\linewidth}
\centering
\includegraphics[width=1\textwidth]{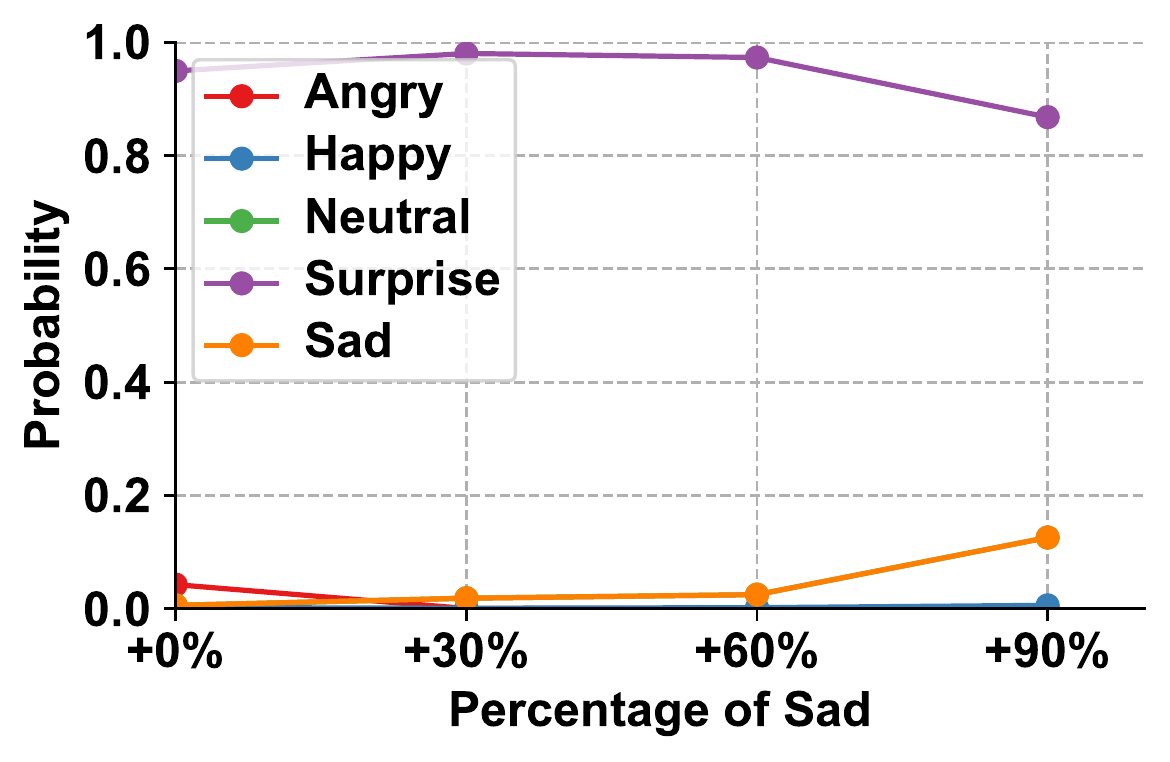}
\end{minipage}
}
\centering
\caption{Classification probabilities derived from the pre-trained SER model for a female speaker ('0019') from the ESD dataset.  Each point represents an averaged probability value of 20 utterances with mixed emotions.}
\label{fig:ser_female}
\end{figure*}

\subsubsection{Training Pipeline}

We first pre-train a relative ranking function between each emotion pair using an emotional speech dataset. We implement the relative ranking function following an open-source repository\footnote{\url{https://github.com/chaitanya100100/Relative-Attributes-Zero-Shot-Learning}}. We use 
a standardized set of 
$384$ acoustic features extracted with openSMILE \cite{eyben2010opensmile} as the input features. These features include zero-crossing rate, frame energy, pitch frequency, and Mel-frequency cepstral coefficient (MFCC) used in the Interspeech Emotion Challenge \cite{schuller2009interspeech}. {The trained ranking functions reported a classification accuracy of $97\%$ on the test set.}

We then conduct a two-stage training strategy to train our text-to-speech framework, which consists of (1) Multi-speaker text-to-speech training with the VCTK Corpus \cite{veaux2016vctk} and (2) Emotion Adaptation for text-to-speech with a single speaker from the ESD dataset \cite{zhou2021seen,zhou2021emotional}. The proposed text-to-speech framework learns abundant speaker styles with a multi-speaker corpus and then learns the actual emotion information with a small amount of emotional speech data. The training strategy we used is similar to that of \cite{zhou21b_interspeech}. During the training, we use the Adam optimizer \cite{kingmaadam} and set the batch size to $64$ and $4$ for multi-speaker text-to-speech training and emotion adaptation, respectively. We set the learning rate to $0.001$ and the weight decay to $0.0001$ for multi-speaker text-to-speech training. We halve the learning rate every seven epochs during the emotion adaptation.

\subsubsection{Data Preparation}
\label{data partition}
We select the VCTK Corpus \cite{veaux2016vctk} to perform multi-speaker text-to-speech training, where we use $99$ speakers and the total duration of training speech data is about $30$ hours. 
We select the ESD dataset \cite{zhou2021seen,zhou2021emotional} to perform emotion adaptation and relative ranking training.  
 We choose one English male ('0013') and one English female ('0019') speaker from the ESD. We consider five emotions: \textit{Neutral}, \textit{Angry}, \textit{Happy}, \textit{Sad} and \textit{Surprise}, and for each emotion, we follow the data partition given in the ESD. For each speaker and each emotion, we use $300$, $30$ and $20$ utterances for training, testing,
and evaluation, respectively. The total duration of emotional speech training data is around $50$ minutes. 

\begin{figure*}[t]
\centering
\subfloat[Mixing Surprise (100\%) with different percentages of Angry (0\%, 30\%, 60\%, 90\%)]
{
\begin{minipage}[c]{0.33\linewidth}
\centering
\includegraphics[width=1\textwidth]{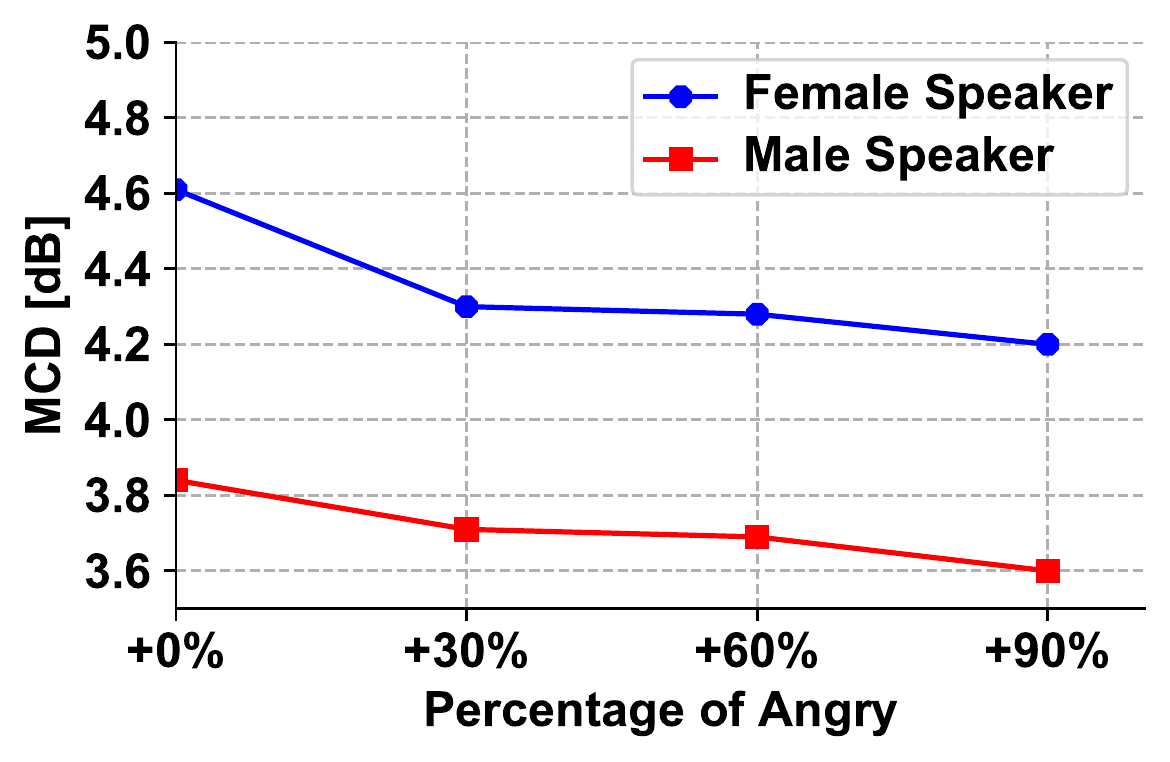}
\end{minipage}%
}
\subfloat[Mixing Surprise (100\%) with different percentages of Happy (0\%, 30\%, 60\%, 90\%)]
{
\begin{minipage}[c]{0.33\linewidth}
\centering
\includegraphics[width=1\textwidth]{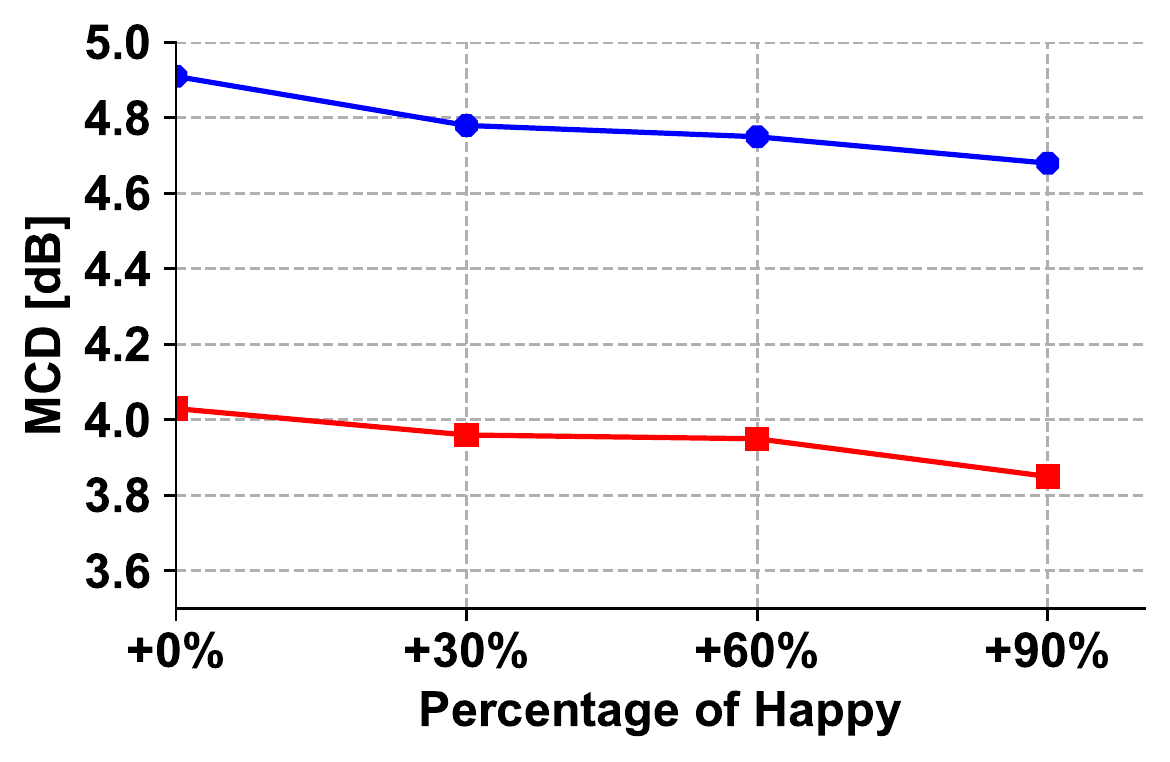}
\end{minipage}
}
\subfloat[Mixing Surprise (100\%) with different percentages of Sad (0\%, 30\%, 60\%, 90\%)]
{
\begin{minipage}[c]{0.33\linewidth}
\centering
\includegraphics[width=1\textwidth]{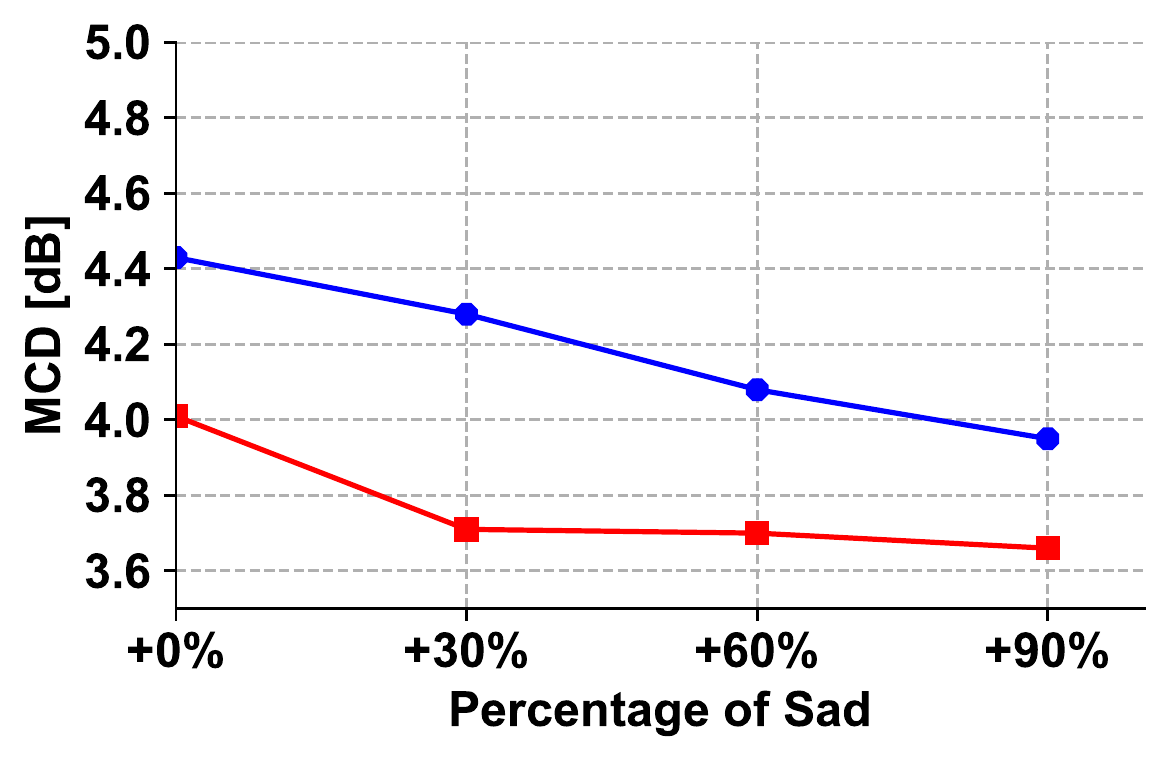}
\end{minipage}
}
\centering
\caption{Mel-cepstral distortion (MCD) [dB] calculated between the Mel-cepstral coefficients (MCEPs) of mixed emotions and the reference emotions (Angry, Happy and Sad). Each point represents an averaged MCD value of 20 utterances with mixed emotions.  }
\label{fig:mcd}
\end{figure*}

\begin{figure*}[t]
\centering
\subfloat[Mixing Surprise (100\%) with different percentages of Angry (0\%, 30\%, 60\%, 90\%)]
{
\begin{minipage}[c]{0.33\linewidth}
\centering
\includegraphics[width=1\textwidth]{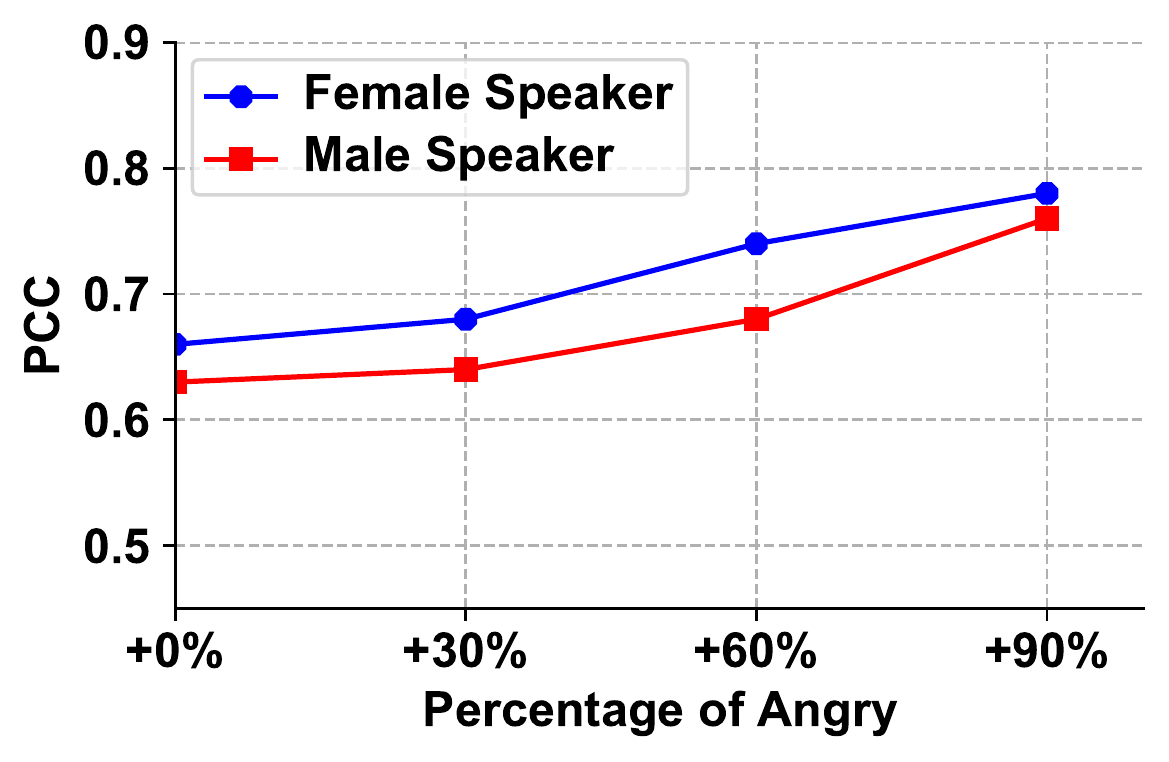}
\end{minipage}%
}
\subfloat[Mixing Surprise (100\%) with different percentages of Happy (0\%, 30\%, 60\%, 90\%)]
{
\begin{minipage}[c]{0.33\linewidth}
\centering
\includegraphics[width=1\textwidth]{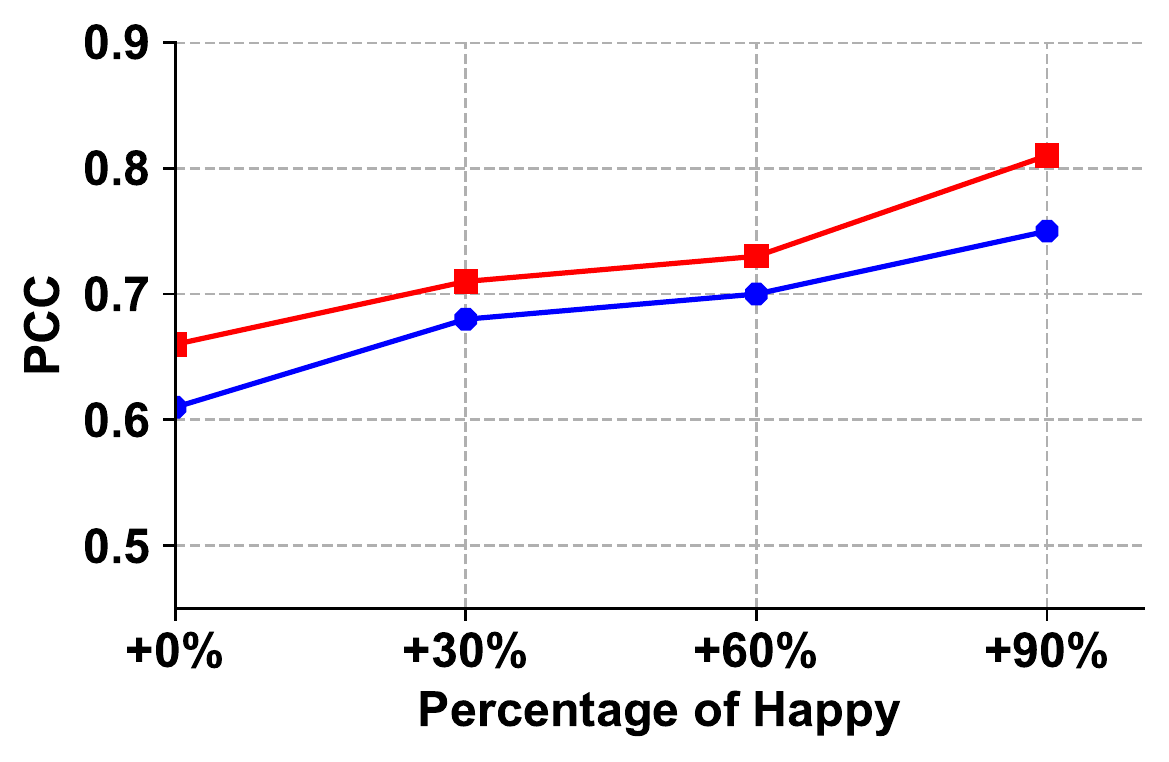}
\end{minipage}
}
\subfloat[Mixing Surprise (100\%) with different percentages of Sad (0\%, 30\%, 60\%, 90\%)]
{
\begin{minipage}[c]{0.33\linewidth}
\centering
\includegraphics[width=1\textwidth]{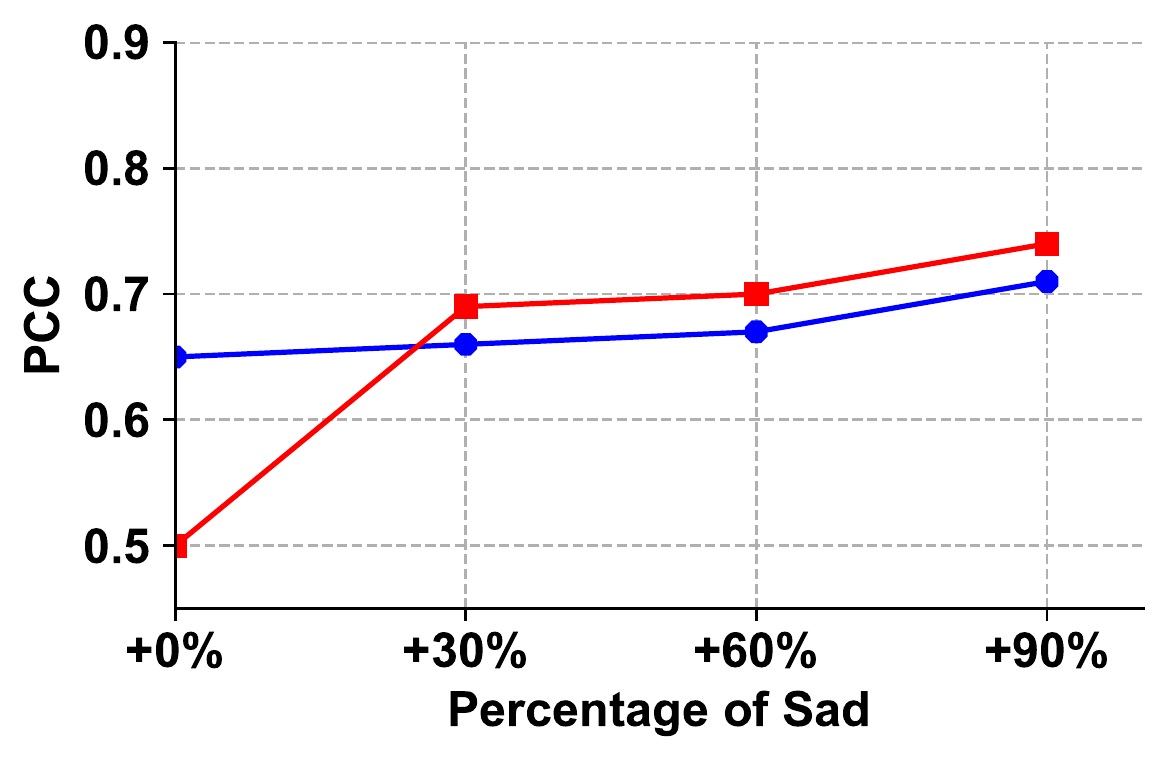}
\end{minipage}
}
\centering
\caption{Pearson Correlation Coefficient (PCC) calculated between the fundamental frequency (F0) of mixed emotions and the reference emotions (Angry, Happy and Sad). Each point represents an averaged PCC value of 20 utterances with mixed emotions.}
\label{fig:pcc}
\end{figure*}

\subsection{Objective Evaluation}
We first perform objective evaluations to validate the proposed mixed emotion synthesis. We demonstrate the effectiveness of our proposals and provide analysis with a pre-trained speech emotion recognition (SER) model. We calculate Mel-cepstral distortion (MCD) and Pearson correlation coefficient (PCC) as objective evaluation metrics.

\subsubsection{Analysis with Speech Emotion Recognition}


{We train a speech emotion recognition model on the ESD dataset \cite{zhou2021seen} with the same data partition described in Section \ref{data partition}.
To improve the robustness of SER, data augmentation is performed by adding white Gaussian noise during the SER training \cite{heracleous2017speech,tiwari2020multi,abbaschian2021deep,muthusamy2015improved}.}

The SER architecture is the same as that in \cite{chen20183}, which includes: 1) a three-dimensional (3-D) CNN layer; 2) a BLSTM; 3) an attention layer; and 4) a fully connected (FC) layer. We evaluate our synthesized mixed emotions with the pre-trained SER. We use the classification probabilities derived from the softmax layer of the SER to analyze the effects of mixed emotions. As a high-level feature, the classification probabilities summarize the useful emotion information from the previous layers for final decision-making. 
{The classification probabilities offer us an effective tool to justify how well each emotional component can be perceptually recognized by the SER from the emotion mixture.}

We first report the classification probabilities for a male speaker ('0013') in Figure \ref{fig:ser_male}. We evaluate four different combinations where we gradually increase the percentage (0\%, 30\%, 60\%, 90\%) of \textit{Angry}, \textit{Happy} or \textit{Sad} while keeping that of \textit{Surprise} always being 100\%. As shown in Figure \ref{fig:ser_male}(a), we observe that the probability of \textit{Angry}
increases while we increase the percentage of \textit{Angry} from 0\% to 90\%. In the meanwhile, the probability of \textit{Surprise} decreases but still remains to be higher than for others. The probability of \textit{Angry} achieves 0.25 when the percentage of \textit{Angry} reaches 90\%. We also note similar observations for \textit{Happy} and \textit{Sad} as shown in Figure \ref{fig:ser_male}(b) and (c).

We then report the classification probabilities for a female speaker ('0019') in Figure \ref{fig:ser_female}. Similar to that of the male speaker, we report four different percentages (0\%, 30\%, 60\%, 90\%) of \textit{Angry}, \textit{Happy} or \textit{Sad} while keeping that of \textit{Surprise} being 100\%. For \textit{Happy}, we observe the probability of \textit{Happy} considerably increases while we increase the percentage of \textit{Happy} in mixed emotions as shown in Figure \ref{fig:ser_female}(b). For \textit{Angry} and \textit{Sad}, we find similar observations as in Figure \ref{fig:ser_female}(a) and (c). These observations indicate that the mixed emotions can be perceptually recognized by a pre-trained SER.

\subsubsection{Mel-cepstral Distorion}
Spectral features, based on the short-term power spectrum of sound, such as Mel-cepstral coefficients (MCEPs), contain rich information about expressivity and emotion \cite{bitouk2010class}.
Mel-cepstral Distortion (MCD) \cite{kubichek1993mel} is a widely adopted metric to measure the spectrum similarity, which is calculated between the synthesized ($\hat{\mathbf{y}}=\{\hat{\mathbf{y}}_m\}$) and the target MCEPs (${\mathbf{y}}=\{\mathbf{{y}}_m\}$):
\begin{equation}
    \text{MCD [dB]} = \frac{10\sqrt{2}}{\ln 10}\frac{1}{M}\sqrt{\sum_{m=1}^M(\mathbf{y_{m}} - \hat{\mathbf{y}}_{m})^2}, 
\end{equation}
where $M$ represents the dimension of the MCEPs. A lower value of MCD indicates a higher degree of in the spectrum.

\subsubsection{Pearson Correlation Coefficient}
Pitch is considered a major prosodic factor contributing to speech emotion, closely correlated to the activity level \cite{johnson1986recognition,owren2007measuring}. In practice, the pitch is often represented by the fundamental frequency (F0), which can be estimated with the harvest algorithm \cite{morise2017harvest}. 
We calculate the Pearson Correlation Coefficient (PCC) of F0 to measure the linear dependency between two F0 sequences, which has been used in previous studies \cite{zhou2020converting, Zhou2020,zhou2021vaw}.
The PCC between two F0 sequences is given as:
\begin{equation}
    \rho(F_0^s,F_0^t) = \frac{cov(F_0^s,F_0^t)}{\sigma_{F_0^s}\sigma_{F_0^t}}, 
\end{equation}
where $cov(\cdot)$ represents the covariance function, $\sigma_{F_0^s}$ and $\sigma_{F_0^t}$ are the standard deviations of the synthesized sequences ($F_0^s$) and the target F0 sequences ($F_0^t$), respectively. A higher PCC value represents a higher degree of similarity in prosody. 

\subsubsection{Discussion of the MCD and PCC Results}
To show the effectiveness of synthesizing mixed emotions, we calculate MCD and PCC between the synthesized results and the reference emotions (\textit{Angry}, \textit{Happy} and \textit{Sad}). We choose one male ('0013') and one female speaker ('0019') from the ESD dataset \cite{zhou2021seen}. For each speaker, we use 20 utterances for evaluation.  We report four different percentages of \textit{Angry}, \textit{Happy} and \textit{Sad} that are: 0\%, 30\%, 60\% and 90\%. Again, we keep \textit{Surprise} as the primary emotion that has  a percentage of \textit{Surprise} is always 100\%.

We first compare spectrum similarity as shown in Figure \ref{fig:mcd}. For all three different combinations, we observe that the MCD values decrease as the percentage of reference emotions (\textit{Angry}, \textit{Happy} and \textit{Sad}) increases as shown in Figure \ref{fig:mcd}(a), (b) and (c). These results show that the synthesized emotion becomes more similar to the reference emotions in the spectrum as we increase the percentage of the reference emotions. 

We have similar observations for prosody similarity as shown in Figure \ref{fig:pcc}. As the percentage of reference emotions (\textit{Angry}, \textit{Happy} and \textit{Sad}) increases, we observe that the PCC value consistently increases. It indicates that the synthesized mixed emotions have a stronger correlation with the reference emotions (\textit{Angry}, \textit{Happy} and \textit{Sad}) in terms of the prosody variance. These results show that we can effectively synthesize and further control the rendering of mixed emotions in terms of the spectrum and prosody. 

\begin{table}[]
\caption{Mean Opinion Score (MOS) with 95\% confidence interval to evaluate the speech quality of synthesized mixed emotions.}
\begin{tabular}{cc|c}
\toprule
\multicolumn{2}{c}{\textbf{Configuration}}                                                                                                         & \textbf{MOS}  \\ \midrule
\multicolumn{2}{c}{Ground truth (Surprise)}                                                                                               & 4.83 $\pm$ 0.16 \\ \midrule
\multicolumn{1}{c|}{\multirow{5}{*}{\begin{tabular}[c]{@{}c@{}}Mixing Surprise (100\%) \\\\ with Angry\end{tabular}}} & Ground truth (Angry) & 4.81 $\pm$ 0.19 \\ 
\multicolumn{1}{c|}{}                                                                                               & + 0\% Angry          & 3.51 $\pm$ 0.36 \\ 
\multicolumn{1}{c|}{}                                                                                               & + 30\% Angry         & 3.79 $\pm$ 0.37 \\ 
\multicolumn{1}{c|}{}                                                                                               & + 60\% Angry         & 3.81 $\pm$ 0.35 \\ 
\multicolumn{1}{c|}{}                                                                                               & + 90\% Angry         & 3.76 $\pm$ 0.35 \\ \midrule
\multicolumn{1}{c|}{\multirow{5}{*}{\begin{tabular}[c]{@{}c@{}}Mixing Surprise (100\%) \\\\ with Happy\end{tabular}}} & Ground truth (Happy) & 4.93 $\pm$ 0.05 \\ 
\multicolumn{1}{c|}{}                                                                                               & + 0\% Happy          & 3.21 $\pm$ 0.41 \\ 
\multicolumn{1}{c|}{}                                                                                               & + 30\% Happy         & 3.36 $\pm$ 0.36 \\ 
\multicolumn{1}{c|}{}                                                                                               & + 60\% Happy         & 3.39 $\pm$ 0.39 \\ 
\multicolumn{1}{c|}{}                                                                                               & + 90\% Happy         & 3.52 $\pm$ 0.42 \\ \midrule
\multicolumn{1}{c|}{\multirow{5}{*}{\begin{tabular}[c]{@{}c@{}}Mixing Surprise (100\%) \\\\ with Sad\end{tabular}}}   & Ground truth (Sad)   & 4.84 $\pm$ 0.15 \\ 
\multicolumn{1}{c|}{}                                                                                               & + 0\% Sad            & 3.64 $\pm$ 0.35 \\ 
\multicolumn{1}{c|}{}                                                                                               & + 30\% Sad           & 3.73 $\pm$ 0.32 \\ 
\multicolumn{1}{c|}{}                                                                                               & + 60\% Sad           & 3.74 $\pm$ 0.31 \\ 
\multicolumn{1}{c|}{}                                                                                               & + 90\% Sad           & 3.60 $\pm$ 0.38  \\ \bottomrule
\end{tabular}
\label{tab: mos}
\end{table}

\subsection{Subjective Evaluation}
We conduct subjective evaluations with human listeners, whom we ask to focus on two aspects: (1) Speech Quality and (2) Emotion Perception. 

\subsubsection{Speech Quality}

We first conduct the Mean Opinion Score (MOS) test to evaluate speech quality, covering the speech's naturalness, intelligibility and listening efforts. All participants are asked to listen to the reference speech (``Ground truth'') and the synthesized speech with mixed emotions and score the ``quality'' of each speech sample on a 5-point scale (`5' for excellent, `4' for good, `3' for fair, `2' for poor, and `1' for bad). $20$ subjects 
listened to $80$ speech samples 
in total ($80$ = $5$ x $4$ (\# of percentages) x $3$ (\textit{Angry}, \textit{Happy} and \textit{Sad}) + $20$ (\# of Ground truth)). 
The actual speech samples can be found in our demo website. We report the MOS results in Table \ref{tab: mos}, which show that our synthesized mixed emotions retain the speech quality between fair and good.

\begin{table}[t]
\caption{Best-worst scaling (BWS) test results to evaluate the perception of the reference emotions (\textit{Angry}, \textit{Happy}, and \textit{Sad}) in synthesized mixed emotions.}
\subfloat[Perception of \textit{Angry}\label{tab:bws1:a}]{
    \centering
\begin{tabular}{cl|l|l}
\toprule
\multicolumn{2}{c|}{\textbf{Configuration}}                                                                                                 & \textbf{Best (\%)}   & \textbf{Worst (\%)}  \\ \midrule
\multicolumn{1}{c|}{\multirow{4}{*}{\begin{tabular}[c]{@{}c@{}}Mixing Surprise (100\%)\\with Angry\end{tabular}}} & + 0\% Angry  & 8.3  & 61.7 \\ 
\multicolumn{1}{c|}{}                                                                                               & + 30\% Angry & 6.0  & 19.5 \\ 
\multicolumn{1}{c|}{}                                                                                               & + 60\% Angry & 24.8 & 11.3 \\ 
\multicolumn{1}{c|}{}                                                                                               & + 90\% Angry & \textbf{60.9} & \textbf{7.5}  \\ \bottomrule
\end{tabular}}\\
\subfloat[Perception of \textit{Happy}\label{tab:bws1:b}]{
    \centering
\begin{tabular}{cl|l|l}
\toprule
\multicolumn{2}{c|}{\textbf{Configuration}}                                                                                                 & \textbf{Best (\%)}   & \textbf{Worst (\%)}  \\ \midrule
\multicolumn{1}{c|}{\multirow{4}{*}{\begin{tabular}[c]{@{}c@{}}Mixing Surprise (100\%)\\with Happy\end{tabular}}} & + 0\% Happy  & 8.3  & 44.4 \\ 
\multicolumn{1}{c|}{}                                                                                               & + 30\% Happy & 24.0  & 25.6 \\ 
\multicolumn{1}{c|}{}                                                                                               & + 60\% Happy & 27.1 & \textbf{11.2} \\ 
\multicolumn{1}{c|}{}                                                                                               & + 90\% Happy & \textbf{40.6} & 18.8  \\ \bottomrule
\end{tabular}}\\
\subfloat[Perception of \textit{Sad}\label{tab:bws1:c}]{
    \centering
\begin{tabular}{cl|l|l}
\toprule
\multicolumn{2}{c|}{\textbf{Configuration}}                                                                                                 & \textbf{Best (\%)}   & \textbf{Worst (\%)}  \\ \midrule
\multicolumn{1}{c|}{\multirow{4}{*}{\begin{tabular}[c]{@{}c@{}}Mixing Surprise (100\%)\\with Sad\end{tabular}}} & + 0\% Sad  & 9.0  & 57.1 \\ 
\multicolumn{1}{c|}{}                                                                                               & + 30\% Sad & 9.0  & 29.3 \\ 
\multicolumn{1}{c|}{}                                                                                               & + 60\% Sad & 20.3 & \textbf{3.8} \\ 
\multicolumn{1}{c|}{}                                                                                               & + 90\% Sad & \textbf{61.7} & 9.8  \\ \bottomrule
\end{tabular}}
\label{tab: bws1}
\end{table}

\begin{table}[t]
\caption{Best-worst scaling (BWS) test results to evaluate the perception of mixed emotional effects (\textit{Outrage}, \textit{Delight}, and \textit{Disappointment}) in synthesized mixed emotions.}
\subfloat[Perception of \textit{Outrage}\label{tab:bws2:a}]{
    \centering
\begin{tabular}{cl|l|l}
\toprule
\multicolumn{2}{c|}{\textbf{Configuration}}                                                                                                 & \textbf{Best (\%)}   & \textbf{Worst (\%)}  \\ \midrule
\multicolumn{1}{c|}{\multirow{4}{*}{\begin{tabular}[c]{@{}c@{}}Mixing Surprise (100\%)\\with Angry\end{tabular}}} & + 0\% Angry  & 6.8  & 61.7 \\ 
\multicolumn{1}{c|}{}                                                                                               & + 30\% Angry & 4.5  & 23.3 \\ 
\multicolumn{1}{c|}{}                                                                                               & + 60\% Angry & 15.8 & 9.8 \\ 
\multicolumn{1}{c|}{}                                                                                               & + 90\% Angry & \textbf{72.9} & \textbf{5.2}  \\ \bottomrule
\end{tabular}}\\
\subfloat[Perception of \textit{Delight}\label{tab:bws2:b}]{
    \centering
\begin{tabular}{cl|l|l}
\toprule
\multicolumn{2}{c|}{\textbf{Configuration}}                                                                                                 & \textbf{Best (\%)}   & \textbf{Worst (\%)}  \\ \midrule
\multicolumn{1}{c|}{\multirow{4}{*}{\begin{tabular}[c]{@{}c@{}}Mixing Surprise (100\%)\\with Happy\end{tabular}}} & + 0\% Happy  & 5.3  & 66.2 \\ 
\multicolumn{1}{c|}{}                                                                                               & + 30\% Happy & 10.5  & 21.8 \\ 
\multicolumn{1}{c|}{}                                                                                               & + 60\% Happy & 30.1 & \textbf{3.0} \\ 
\multicolumn{1}{c|}{}                                                                                               & + 90\% Happy & \textbf{54.1} & 9.0  \\ \bottomrule
\end{tabular}}\\
\subfloat[Perception of \textit{Disappointment}\label{tab:bws2:c}]{
    \centering
\begin{tabular}{cl|l|l}
\toprule
\multicolumn{2}{c|}{\textbf{Configuration}}                                                                                                 & \textbf{Best (\%)}   & \textbf{Worst (\%)}  \\ \midrule
\multicolumn{1}{c|}{\multirow{4}{*}{\begin{tabular}[c]{@{}c@{}}Mixing Surprise (100\%)\\with Sad\end{tabular}}} & + 0\% Sad  & 11.3  & 54.9 \\ 
\multicolumn{1}{c|}{}                                                                                               & + 30\% Sad & 12.0  & 26.4 \\ 
\multicolumn{1}{c|}{}                                                                                               & + 60\% Sad & 14.3 & \textbf{9.0} \\ 
\multicolumn{1}{c|}{}                                                                                               & + 90\% Sad & \textbf{62.4} & 9.7  \\ \bottomrule
\end{tabular}}
\label{tab: bws2}
\end{table}

\subsubsection{Emotion Perception}
\label{perception}

We then conduct the best-worst scaling (BWS) test to evaluate the emotion perception of synthesized mixed emotions. All participants are asked to listen to the speech samples and choose the best and the worst one according to their perception of a specific emotion type. $20$ subjects listened to $168$ speech samples in total (168 = 7 x 4 (\# of percentages) x 6 (\textit{Angry}, \textit{Happy}, \textit{Sad}, \textit{Outrage}, \textit{Delight} and \textit{Disappointment})). The actual speech samples can be found on our demo website.

We first evaluate the perception of the reference emotions (\textit{Angry}, \textit{Happy} and \textit{Sad}) that are mixed with \textit{Surprise}. As shown in Table \ref{tab:bws1:a}, \ref{tab:bws1:b} and \ref{tab:bws1:c}, the mixed emotion with 90\% of the reference emotions consistently achieves the highest percentage of the ``Best'' score; also, the ``Best'' score increases as the percentage of reference emotion increases. Similarly, the highest ``Worst'' score is observed when the reference emotion is added at the lowest percentage (0\%). These results confirm the effectiveness of controlling the rendering of mixed emotions. {We also observe a slight rise of the worst rating  when the percentage of \textit{Happy} and \textit{Sad} exceeds 60\% in Table \ref{tab:bws1:b}, and \ref{tab:bws1:c}. 
This observation we attribute to the unnatural emotional expressions that may be created to influence listeners' preferences.} 

We then take one step further to evaluate the perception of \textit{Outrage}, \textit{Delight} and \textit{Disappointment} in synthesized speech. In psychology, there is evidence that those feelings could be produced by combining several emotions. We observe that participants can perceive such feelings, and most of them choose those with 90\% of reference emotions as the ``Best'', as shown in Table \ref{tab:bws2:a}, \ref{tab:bws2:b} and \ref{tab:bws2:c}. {As for the rating of ``Worst", we also have similar observations to those in Table \ref{tab: bws1}.} These results show that we can synthesize new emotion types that are subtle and hard to collect in real life, which will significantly benefit the research community.

\subsection{Ablation Study}

We further conduct ablations studies to validate the contributions of the proposed relative scheme on emotional expression. We compare the proposed framework with or without the relative scheme through several XAB preference tests, where the participants are asked to listen to the reference emotional speech first, then choose the one closer to the reference in terms of emotional expression. 20 subjects 
listened to 60 speech samples in total (60 = 5 x 2 (\# of frameworks) x 4 (\# of emotions) + 20 (\# of ground truth)). 

\begin{figure}[t]
    \centering
    \includegraphics[width=0.48\textwidth]{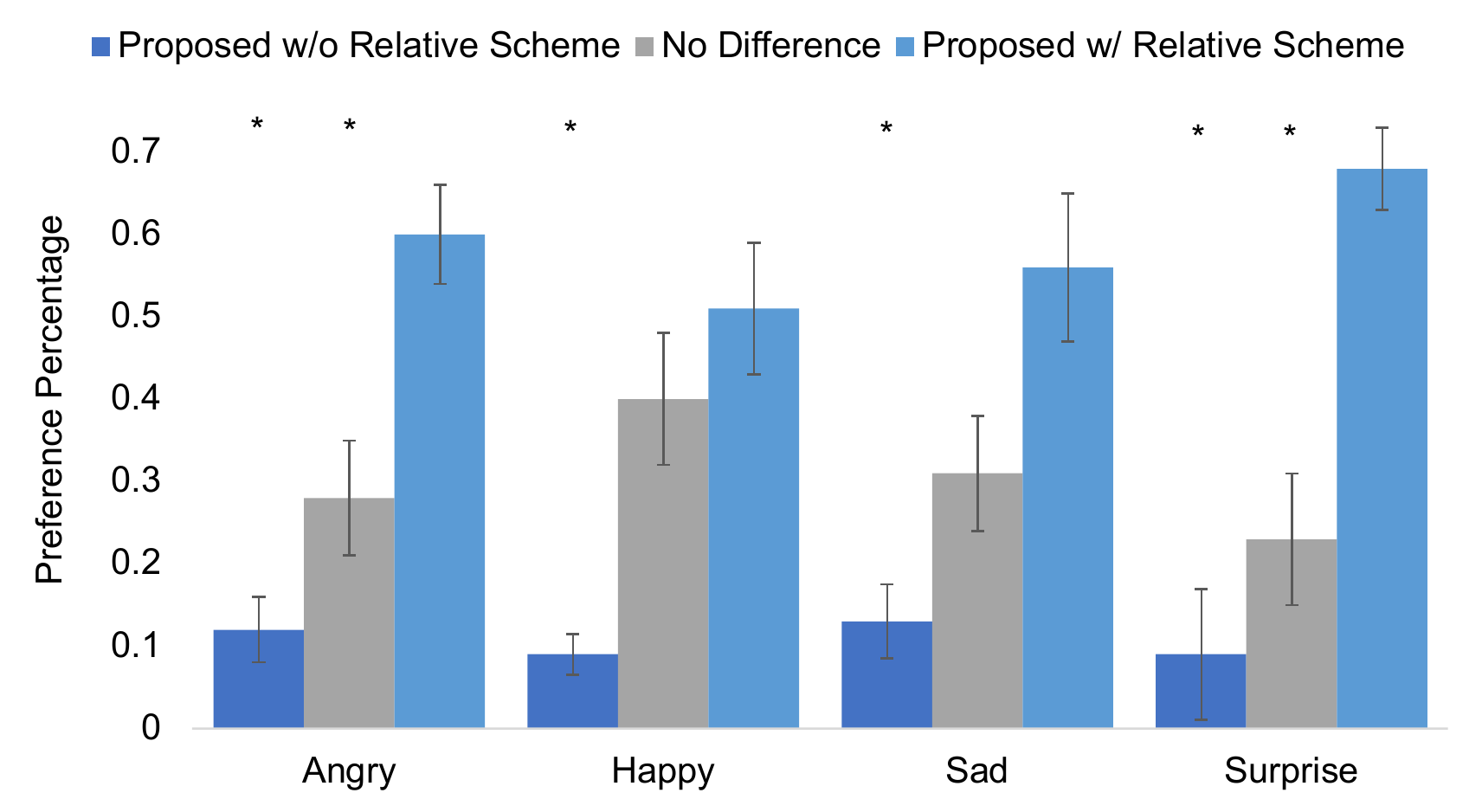}
    \caption{XAB preference test results with 95\% confidence interval to evaluate the emotion similarity with the ground truth emotions.  The marker * indicates p $<$ 0.05 for paired t-test scores (pairs between ``Proposed w/ Relative Scheme'' and the others).}
    \label{fig:xab}
\end{figure}

We report the XAB results in Figure \ref{fig:xab} where we observe that ``Proposed w/ Relative Scheme'' consistently and considerably outperforms ``Proposed w/o Relative Scheme'' for all emotions (\textit{Angry}, \textit{Happy}, \textit{Sad} and \textit{Surprise}). Besides, the $p$ values calculated between those two pairs (``Proposed w/ Relative Scheme'' and ``Proposed w/o Relative Scheme'') are always lower than $0.05$, indicating that the out-performance did not occur by chance. These results demonstrate that our relative scheme can improve emotional intelligibility in synthesized emotional speech.

\section{Further Investigations and Discussion}
\label{sec: further}
In this section, we expand our experiments and show the ability of our proposed methods on other interesting topics. We first investigate the mixed emotional effects of \textit{Happy} and \textit{Sad}, which are two oppositely valenced emotions. We then build an emotion transition system with our proposed method. We do not seek to conduct comprehensive evaluations but to provide some interesting insights into mixed emotion synthesis and its applications. All the speech samples are provided on the demo page.

\subsection{Oppositely Valenced Emotions: Happy and Sad}
In our experiments, we mostly focus on mixing \textit{Surprise} with other emotions (\textit{Angry}, \textit{Happy} and \textit{Sad}), which is thought to be easier to perceive for human listeners. Here, we move one step further to study a more challenging task, which is to synthesize mixed effects of \textit{Happy} and \textit{Sad}. In Russell's valence-arousal model \cite{russell1980circumplex}, \textit{Happy} and \textit{Sad} are two conflicting emotions with opposite valance (\textit{Pleasant} and \textit{Unpleasant}). There are some debates that agree with the co-existence of conflicting emotions \cite{williams2002can,miyamoto2010culture}. In real life, there are also some terms to describe such feelings in different cultures, for example, ``\textit{Bittersweet}'' in English.
Professional actors are thought to be able to deliver such feelings to the audience through both actions and speech.
With our proposed methods, we are able to synthesize such mixed feelings of the oppositely valenced emotions such as \textit{Happy} and \textit{Sad}. Readers are suggested to refer to the demo page. 
\subsection{An Emotion Transition System}

One potential application of mixed emotion synthesis is building an emotion transition system \cite{hareli2016role}. Emotion transition aims to gradually transition the emotion state from one to another. One similar study is emotional voice conversion \cite{zhou2021emotional}, which aims to convert the emotional state. 
Compared with emotional voice conversion, 
the key challenge of emotion transition is to synthesize internal states between different emotion types. With our proposed methods, we are able to model these internal states by mixing them with different emotions. To achieve this, the sum up of the percentages of each emotion needs to be 100\% (e.\,g., 80\% \textit{Surprise} with 20\% \textit{Angry}; 40\% \textit{Happy} with 60\% \textit{Sad}). Then, we can synthesize various internal emotion states by adjusting the percentages. 
Compared with traditional methods such as interpolation, our proposed system is data-driven, and the synthesized emotions are more natural.

\subsection{Discussion}

{This study serves as the first attempt to model and synthesize mixed emotions for speech synthesis. Although we have shown the effectiveness of our methods, the related problems have not been completely solved. We provide a discussion to address the concerns, show our findings, and inspire future studies.}

\subsubsection{Category vs. Dimensional Emotion Models}

{Our assumptions, formulation, and evaluation of mixed emotions are all based on categorical emotion studies. 
We note that mixed emotions can also be modelled with dimensional representations such as arousal, valence, and dominance. A dimensional model can capture a wide range of emotional concepts, which offers a means of measuring the similarity of different emotional states \cite{ps2017emotion}. However, several problems need to be adequately dealt with when modelling mixed emotions with a dimensional model. As mentioned in Section \ref{characterization}, the significant challenge for using dimensional representations comes from the lack of labels. Besides, humans are more efficient at discriminating among options than giving an absolute score \cite{miller1956magical}, which adds challenges to the evaluation process. Furthermore, dimensional models are restricted to modelling the co-occurrence of like-valenced discrete emotions \cite{barrett1998discrete}. For these reasons, we refrain from applying dimensional emotions to the current framework.}

\subsubsection{Remaining Challenges}

{There are a few remaining challenges that need attention from the community. As mentioned in Section \ref{perception}, increasing the percentage of adding emotions may result in unnatural emotional expressions. If the synthesized emotion sounds unnatural or is difficult to understand, it may not be effective in achieving the desired outcome. Additionally, the human voice is a complex and highly variable instrument, and different people can produce the same emotional state in very different ways. This can make it difficult to accurately capture and reproduce a desired mix of emotions. At last, human raters are asked to evaluate the mixed emotions totally based on their personal experiences because of the lack of ``ground truth" emotions. People from different cultures may have different experiences and backgrounds that can influence their emotional responses, and having a diverse group of evaluators can provide a more well-rounded perspective on the synthesized emotions.}

\subsubsection{Potential Improvements}
{We discuss several potential improvements to inspire future studies on mixed emotion synthesis: 1) Selection of ranking functions: adopt deep learning-based ranking methods \cite{koch2015siamese} to improve the performance of ranking; 2) Multi-speaker studies: add training data from multiple speakers; 3) Non-autoregressive backbone frameworks: use non-autoregressive TTS framework as the backbone to avoid the misalignment of attention and improve the naturalness of synthesized speech.}
\section{Conclusion}
\label{sec: concludes}

This contribution fills the gap on mixed emotion synthesis in the literature on speech synthesis. We proposed an emotional speech synthesis framework that is based on a sequence-to-sequence model. For the first time, with the proposed framework, we are able to synthesize mixed emotions and further control the rendering of mixed emotions at run-time.
The key highlights are as follows:

\begin{enumerate}

\item We proposed a novel relative scheme to measure the difference between each emotion pair. We demonstrate that our proposed relative scheme enables the effective synthesis and control of the rendering of mixed emotions. Through ablation studies, we also show that the proposed relative scheme improves emotional intelligibility in synthesized speech;

\item We presented a comprehensive study to evaluate mixed emotions for the first time. Through both objective and subjective evaluations, we validated our idea and showed the effectiveness of our proposed framework in terms of synthesizing mixed emotions; 

\item 
{We present further investigations on synthesising a bittersweet feeling and an emotion triangle. The investigation study serves as an additional contribution to the article, which could broaden the scope of the study.}
\end{enumerate}

In this article, we only focused on studying mixed emotions for emotional text-to-speech. We believe that our proposed relative scheme could enable mixed emotion synthesis in most existing emotional speech synthesis frameworks, including but not limited to emotional text-to-speech. We will expand our experiments to include emotional voice conversion in our future studies. 

{The future work includes: 1) a comparison with other ranking methods such as metric learning \cite{mcfee2010metric} and Siamese neural networks \cite{koch2015siamese}; 2) conducting experiments for more emotion combinations, speakers, and other languages.   }
Our future directions also include the study of cross-lingual emotion style modeling and transfer. 
Besides, a closer look at linguistic prosody for emotional speech synthesis is foreseen; for example, different semantic meanings can affect the way of expressing an emotion.

\ifCLASSOPTIONcompsoc
  \section*{Acknowledgments}
\else
  \section*{Acknowledgment}
\fi

The research by Kun Zhou is supported by the Science and Engineering Research Council, Agency of Science, Technology and Research (A*STAR), Singapore, through the National Robotics Program under Human-Robot Interaction Phase 1 (Grant No. 192 25 00054); Human-Robot Collaborative AI under its AME Programmatic Funding Scheme (Project No.\ A18A2b0046); and the National Research Foundation Singapore under its AI Singapore Programme (Award Grant No. AISG-100E-2018-006). The work by Haizhou Li is supported by the Guangdong Provincial Key Laboratory of Big
Data Computing, The Chinese University of Hong Kong, Shenzhen (Grant
No. B10120210117-KP02), the Research
Foundation of Guangdong Province (Grant No. 2019A050505001) and the National Natural Science Foundation of China (Grant No. 62271432).


\ifCLASSOPTIONcaptionsoff
  \newpage
\fi



%
\newpage
\bibliographystyle{IEEEtran}
\bibliography{Bibliography}

\begin{IEEEbiography}[{\includegraphics[width=1in,height=1.25in,clip,keepaspectratio]{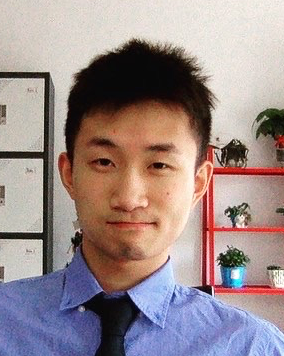}}]{Kun Zhou}(Student Member, IEEE)
received his B.\,Eng.\ degree from the School of Information and Communication Engineering, University of Electronic Science and Technology of China (UESTC), Chengdu, China, in 2018, and the M.\,Sc.\ degree from the Department of Electrical and Computer Engineering, National University of Singapore (NUS), Singapore, in 2019. He is currently a PhD student at the National University of Singapore. He was a visiting PhD student at The Center for Robust Speech Systems (CRSS), the University of Texas at Dallas, United States (2022).
His research interests mainly focus on emotion analysis and synthesis in speech, including emotional voice conversion and emotional text-to-speech. He is the recipient of the PREMIA best student paper award 2022. He served on the organizing committee for IEEE ASRU 2019, SIGDIAL 2021, IWSDS 2021, O-COCOSDA 2021 and ICASSP 2022. He is a reviewer for multiple leading conferences and journals including INTERSPEECH, ICASSP, IEEE SLT, IEEE Signal Processing Letters, Speech Communication and IEEE/ACM Transactions on Audio, Speech and Language Processing.
\end{IEEEbiography}
\vskip -2\baselineskip plus -1fil
\begin{IEEEbiography}[{\includegraphics[width=1in,height=1.25in,clip,keepaspectratio]{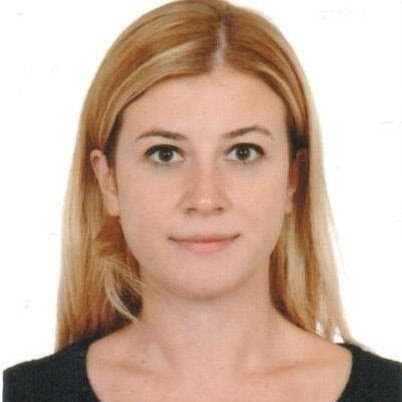}}]{Berrak Sisman}(Member, IEEE)
received a PhD degree in Electrical and Computer Engineering from the National University of Singapore in 2020, fully funded by A*STAR Graduate Academy under Singapore International Graduate Award (SINGA). She is currently working as a tenure-track Assistant Professor at the Erik Jonsson School Department of Electrical and Computer Engineering at the University of Texas at Dallas, United States. Prior to joining UT Dallas, she was a tenure-track faculty at the Singapore University of Technology and Design (2020-2022). She was a Postdoctoral Research Fellow at the National University of Singapore (2019-2020), and a Visiting Researcher at Columbia University, New York, United States (2020). She was an exchange PhD student at the University of Edinburgh and a visiting scholar at The Centre for Speech Technology Research (CSTR), University of Edinburgh (2019). She was attached to RIKEN Advanced Intelligence Project, Japan (2018). Her research is focused on machine learning, signal processing, emotion, speech synthesis and voice conversion. She plays leadership roles in conference organizations and is also active in technical committees. She has served as the General Coordinator of the Student Advisory Committee (SAC) of the International Speech Communication Association (ISCA). She has served as the Area
Chair at INTERSPEECH 2021, INTERSPEECH 2022, IEEE SLT 2022 and as the Publication Chair at ICASSP 2022. She has been elected as a member of the IEEE Speech and Language Processing Technical Committee (SLTC) in the area of Speech Synthesis for the term from Jan. 2022 to Dec. 2024.
\end{IEEEbiography}
\vskip -2\baselineskip plus -1fil

\begin{IEEEbiography}[{\includegraphics[width=1in,height=1.25in,clip,keepaspectratio]{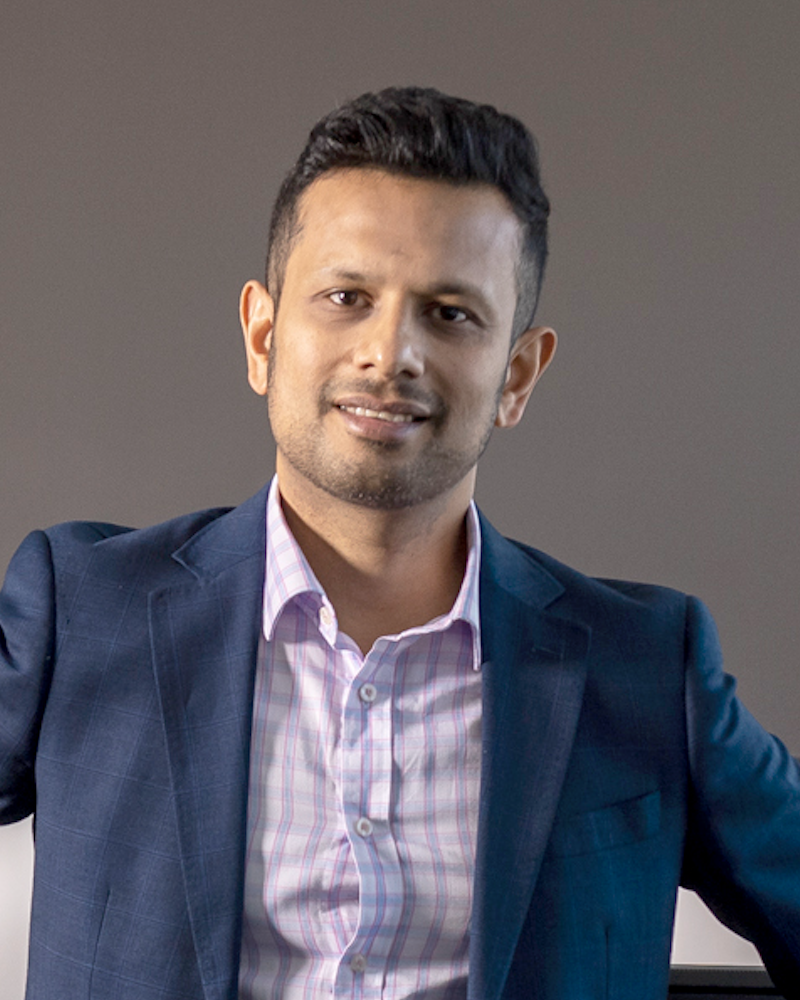}}]{Rajib Rana}(Member, IEEE)
received the B.\,Sc.\ degree in computer science and engineering from Khulna University, with the Prime Minister and President's Gold Medal for outstanding achievements, and the Ph.\,D.\ degree in computer science and engineering from the University of New South Wales, Sydney, Australia, in 2011. He received his Postdoctoral Training with the Autonomous System Laboratory, CSIRO, before joining the University of Southern Queensland, as a Faculty Member, in 2015. 
He is currently a Senior Advance Queensland Research Fellow and an Associate Professor at the University of Southern Queensland. He is also the Director of the 
IoT Health Research Program with the University of Southern Queensland, which capitalizes on advancements in technology and sophisticated information and data processing to understand disease progression in chronic health conditions better and develop predictive algorithms for chronic diseases, such as mental illness and cancer. His current research interests include unsupervised representation learning, Adversarial Machine Learning, Re-enforcement Learning, Federated Learning, Emotional Speech Generation, and Domain Adaptation.
\end{IEEEbiography}
\vskip -2\baselineskip plus -1fil
\begin{IEEEbiography}[{\includegraphics[width=1in,height=1.25in,clip,keepaspectratio]{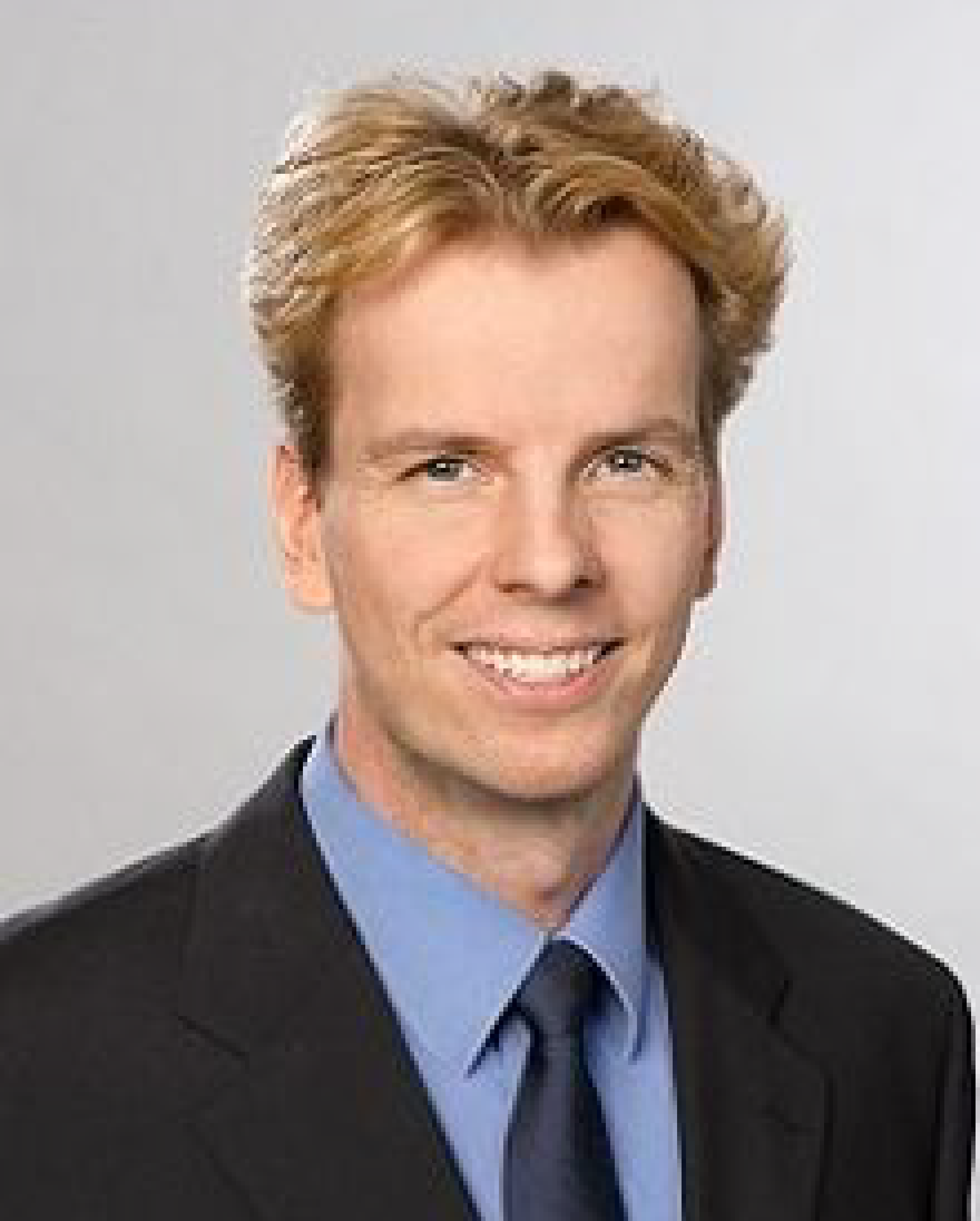}}]{Bj{\"o}rn W. Schuller}(M’05-SM’15-F’18)
received the diploma degree, the doctoral degree in automatic speech and emotion recognition, and the habilitation and Adjunct Teaching Professor in signal processing and machine intelligence from Technische Universit{\"a}t M{\"u}nchen (TUM), Munich, Germany, in 1999, 2006, and 2012, respectively, all in electrical engineering and information technology. He is currently a Professor of Artificial Intelligence with the Department of Computing, Imperial College London, U.K., where he heads the Group on Language, Audio, \& Music (GLAM), a Full Professor and the Head of the Chair of Embedded Intelligence for Health Care and Wellbeing with the University of Augsburg, Germany, and the founding CEO/CSO of audEERING. He was previously a Full Professor and the Head of the Chair of Complex and Intelligent Systems at the University of Passau, Germany. He has (co-)authored five books and more than 1\,000 publications in peer-reviewed books, journals, and conference proceedings leading to more than overall 40,000 citations (H-index=97). He was an Elected Member of the IEEE Speech and Language Processing Technical Committee. He is a Golden Core Member of the IEEE Computer Society, a Fellow of the IEEE, AAAC, BCS, and ISCA, as well as a Senior Member of the ACM, and the President-Emeritus of the Association of the Advancement of Affective Computing (AAAC). He was the General Chair of ACII 2019, a Co-Program Chair of Interspeech, in 2019, and ICMI, in 2019, a repeated Area Chair of ICASSP, next to a multitude of further Associate and Guest Editor roles and functions in Technical and Organisational Committees. He is the Field Chief Editor of the Frontiers in Digital Health and a former Editor-in-Chief of the IEEE Transactions on Affective Computing. 
\end{IEEEbiography}
\vfill
\begin{IEEEbiography}[{\includegraphics[width=1in,height=1.25in,clip,keepaspectratio]{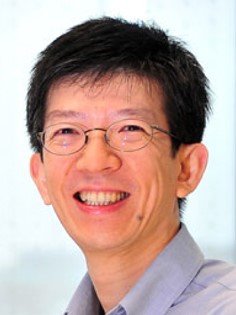}}]{Haizhou Li}
(M’91-SM’01-F’14) received the B.\,Sc., M.\,Sc., and Ph.D degrees in electrical and electronic engineering from South China University of Technology, Guangzhou, China in 1984, 1987 and 1990 respectively. Dr Li is currently a Presidential Chair Professor and the Executive Dean of the School of Data Science, The Chinese University of Hong Kong, Shenzhen, China. He is also with   the Department of Electrical and Computer Engineering, National University of Singapore (NUS). His research interests include automatic speech recognition, speaker and language recognition and natural language processing. Prior to joining NUS, he taught at the University of Hong Kong (1988-1990) and the South China University of Technology (1990-1994). He was a Visiting Professor at CRIN in France (1994-1995), Research Manager at the Apple-ISS Research Centre (1996-1998), Research Director at Lernout \& Hauspie Asia Pacific (1999-2001), Vice President at InfoTalk Corp.\ Ltd.\ (2001-2003) and the Principal Scientist and Department Head of Human Language Technology in the Institute for Infocomm Research, Singapore (2003-2016). Dr Li served as the Editor-in-Chief of IEEE/ACM Transactions on Audio, Speech and Language Processing (2015-2018) and as a Member of the Editorial Board of Computer Speech and Language (2012-2018). He was an elected Member of the IEEE Speech and Language Processing Technical Committee (2013-2015), the President of the International Speech Communication Association (2015-2017), the President of the Asia Pacific Signal and Information Processing Association (2015-2016) and the President of the Asian Federation of Natural Language Processing (2017-2018). He was the General Chair of ACL 2012, INTERSPEECH 2014, ASRU 2019 and ICASSP 2022. Dr Li is a Fellow of the IEEE, the ISCA and the Academy of Engineering Singapore. He was a recipient of the National Infocomm Award in 2002 and the President's Technology Award in 2013 in Singapore. He was named one of the two Nokia Visiting Professors in 2009 by the Nokia Foundation, and Bremen Excellence Chair Professor in 2019.

\end{IEEEbiography}
%

\end{document}